\documentclass[11pt]{article}

\usepackage{microtype}
\usepackage[T1]{fontenc}
\usepackage[utf8]{inputenc}

\usepackage{times}
\usepackage{latexsym}

\usepackage{tabularray}
\usepackage{cellspace}
\usepackage{setspace}
\newcommand\cincludegraphics[2][]{\raisebox{-0.28\height}{\includegraphics[#1]{#2}}}
\usepackage{graphicx}
\usepackage{subcaption}
\usepackage{booktabs} % for professional tables
\usepackage{xspace}
\usepackage{hyperref}

\usepackage{acl}
\usepackage{amsmath}
\usepackage{amssymb}
\usepackage{mathtools}
\usepackage{amsthm}
\usepackage{bbm}
\usepackage{enumitem}
\usepackage{csvsimple}
\usepackage{siunitx} 
\usepackage{ifthen}
\usepackage{adjustbox}

\usepackage{booktabs}
\usepackage{multirow}

\usepackage[capitalize,noabbrev]{cleveref}

\crefname{appendix}{App.}{Apps.}
\crefname{equation}{Eq.}{Eqs.}
\Crefname{appendix}{App.}{Apps.}

\theoremstyle{plain}

\theoremstyle{definition}

\theoremstyle{remark}

 \usepackage[disable]{todonotes}
\newcommand{\note}[4][]{\todo[author=#2,color=#3,size=\scriptsize,fancyline,caption={},#1]{#4}} % default note settings, used by macros below
\newcommand{\tiago}[2][]{\note[#1]{\textbf{Tiago}}{cyan!30}{#2}}
\newcommand{\clara}[2][]{\note[#1]{\textbf{Clara}}{orange!30}{#2}}

\newcommand{\defn}[1]{\textbf{#1}}
\newcommand{\defeq}{\mathrel{\stackrel{\textnormal{\tiny def}}{=}}}
\newcommand{\defpropto}{\mathrel{\stackrel{\textnormal{\tiny def}}{\propto}}}

\newcommand{\mymacro}[2]{\newcommand{#1}{{#2}}}
\mymacro{\unigramdistparams}{\boldsymbol{\phi}}
\mymacro{\unigramdistparamstoken}{\phi_\token}
\mymacro{\unigramdistparamscur}{\unigramdistparams^{(n)}} % current (old)unigram distribution on \vocab
\mymacro{\unigramdistparamscurtoken}{\phi^{(n)}_\token} % current (old)unigram distribution on \vocab
\mymacro{\unigramdistparamsnewtoken}{\phi^{(n+1)}_\token} % current (old)unigram distribution on \vocab
\mymacro{\unigramdistparamslangtoken}{\phi^{\lang}_\token} % current (old)unigram distribution on \vocab

\newcommand{\smalldots}{...}

\mymacro{\unigramdistparamscurminustok}{\unigramdistparams^{(n)}_{-\token}}
\mymacro{\token}{v}
\mymacro{\tokenrv}{V}
\mymacro{\tokenrvuni}{\tokenrv_\unigramdistparams}
\mymacro{\tokenrvcur}{\tokenrv_{n}}
\mymacro{\stringrv}{\mathbf{S}}
\mymacro{\stringrvcur}{\stringrv_{n}}
\mymacro{\tokenseqrv}{\mathbf{V}}
\mymacro{\tokenseqrvcur}{\tokenseqrv_{n}}
\mymacro{\tokenseqrvunilang}{\tokenseqrv_{\unigramdistparamslang}}
\mymacro{\unigramdistparamslang}{\unigramdistparams^\lang}

\mymacro{\vocab}{\mathcal{V}}
\mymacro{\vocabuni}{\vocab_\unigramdistparams}
\mymacro{\stringunit}{s}
\mymacro{\str}{\mathbf{\stringunit}}
\mymacro{\tokens}{\mathbf{\token}}
\mymacro{\stralphabet}{\Sigma}
\mymacro{\resalphabet}{\Gamma}
\mymacro{\stringidx}{t}
\mymacro{\iterationidx}{n}
\mymacro{\strlength}{|\str|}
\mymacro{\toklength}{m}
\mymacro{\tokindex}{t}

\mymacro{\languages}{\Omega}
\mymacro{\lang}{\ell}
\mymacro{\corpus}{\mathcal{C}}
\mymacro{\tokenloss}{\mathrm{loss}}
\mymacro{\likelihood}{\mathcal{L}}
\mymacro{\loss}{L}
\mymacro{\detokfunc}{g}
\mymacro{\tokfunc}{h}

\mymacro{\unigramdistparamsglobal}{\unigramdistparams^{\mathrm{g}}}

\mymacro{\allsegmentations}{\mathcal{T}}
\mymacro{\bestsegmentation}{\tokfunc_{\unigramdistparams}(\str)}
\mymacro{\lengthrv}{M}

\DeclareMathOperator*{\argmax}{\mathrm{argmax}}

\newcommand{\unilm}{UnigramLM\xspace}

\DeclareMathOperator*{\E}{\mathbb{E}}

\newcommand{\pstringparam}[1]{p^S_{#1}}
\newcommand{\pstring}[1]{P(\stringrv\!=\!#1;\unigramdistparams)}
\newcommand{\ptokenseq}[1]{P(\tokenseqrv\!=\!#1;\unigramdistparams)}

\newcommand{\ptokenseqlang}[1]{P(\tokenseqrv\!=\!#1; \unigramdistparamslang)}
\mymacro{\pZ}{\ptokenseq}   % token-sequence distribution
\mymacro{\pX}{\pstring}   % induced string distribution
 
\newcommand{\post}[2]{P(\tokenseqrv\!=\!#1\mid \stringrv\!=\!#2; \unigramdistparams)} % 

\newcommand{\joint}[2]{P(\stringrv\!=\!#1, \tokenseqrv\!=\!#2;\unigramdistparams)} % 

\newcommand{\postcur}[2]{P(\tokenseqrv\!=\!#1\mid \stringrv\!=\!#2;\unigramdistparamscur)} % current posterior over segmentations

\newcommand{\countv}[2]{c_{#1}(#2)}   
\mymacro{\pstringcur}{\pstringparam{\unigramdistparamscur}}   % current distribution over strings
\mymacro{\vocabcur}{\vocab_{\iterationidx}}
\mymacro{\Q}{\mathcal{Q}}                        % EM Q-function
\mymacro{\Hent}{\mathrm{H}}

\title{LangMAP: A Language-Adaptive Approach to Tokenization}

\newcommand{\makesf}[1]{\textsf{{{#1}}}}
\newcommand{\ethemailadress}[1]{\href{mailto:#1@inf.ethz.ch}{\makesf{#1}}}
\newcommand{\camemailadress}[1]{\href{mailto:#1@cam.ac.uk}{\makesf{#1}}}
\newcommand{\epflemailadress}[1]{\href{mailto:#1@epfl.ch}{\makesf{#1}}}

\author{Clara Meister$^1$ \quad Suchir Salhan$^2$ \quad Andrzej Szablewski$^2$ \quad Pietro Lesci$^2$ \\ \textbf{Paula Buttery}$^2$ \quad \textbf{Tiago Pimentel}$^3$ \\
  $^1$EPFL, $^2$University of Cambridge, 
  $^3$ETH Z\"urich\\
   \epflemailadress{clara.meister}\makesf{@epfl.ch},\,
  \makesf{\{\camemailadress{sas245}}, \camemailadress{as3623}, \camemailadress{pl487}, \camemailadress{pjb48}\makesf{\}@cam.ac.uk}, \\
  \ethemailadress{tiago.pimentel}\makesf{@inf.ethz.ch}
  \\
    \begin{tblr}{colspec = {Q[c,m] Q[c,m]}, colsep=10pt, stretch=0}
        \cincludegraphics[width=1.1em, keepaspectratio]{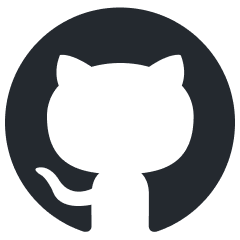} {\fontsize{11pt}{11.5pt}\selectfont\href{https://github.com/cimeister/unimixlm}{cimeister/unimixlm}}
    \end{tblr}
  }

\begin{document}
\maketitle

\providecommand{\crnew}[1]{{\color{red}#1}}

\begin{abstract}

Language-specific tokenizers improve tokenization quality and the downstream performance of models on those languages \citep{rust-etal-2021-how,limisiewicz-etal-2023-tokenization}.
Using one with an existing pretrained model requires either training a new model from scratch or adapting the vocabulary of the pretrained one.
We propose \defn{Language-adaptive Maximum a Posteriori (LangMAP) Tokenization}, which extends the UnigramLM algorithm \citep{kudo-2018-subword} to the multilingual setting by replacing its single distribution over the vocabulary with one distribution per language over the same vocabulary.
LangMAP can be used to train a multilingual language model from scratch, or to adapt a pretrained model's tokenizer to individual languages without changing its vocabulary.
Language labels are required to estimate the per-language distributions; at inference, the algorithm selects the highest-probability segmentation across languages, i.e., it does not need a language label. 
In experiments where we adapt a pretrained tokenizer with LangMAP, we observe that LangMAP's new segmentation has consistently higher morphological boundary recall than the original tokenizer. In code data, it improves alignment with abstract syntax tree (AST) leaf boundaries in all configurations we try.
Despite ostensible improvements in segmentations, in language model fine-tuning experiments, we do not observe significant differences in performance for most language--task pairs.\looseness=-1 
\end{abstract}

\section{Introduction}\label{sec:intro}

Tokenization is a standard pre‑processing step in modern language models (LMs) that converts an input string, a sequence of characters or bytes, into a sequence of tokens from a fixed, finite vocabulary. 
Although tokenization is in principle language‑agnostic, the tokenizers used in most production systems have been optimized for English.
As a result, texts from non-English languages, and in particular those with rich morphology or non-Latin scripts, are tokenized systematically worse: they are over-segmented, producing longer token sequences in which tokens often do not correspond to full words or meaningful morphemes \citep{petrov-etal-2023-tokenizers,lee-etal-2024-length}.
Empirical studies show that these issues hurt downstream performance \citep{lesci-etal-2025-causal} and inflate per‑request costs for commercial APIs, introducing cross-lingual disparities in both the quality and price of available models \citep[][\textit{inter alia}]{ahia-etal-2023-languages, petrov-etal-2023-tokenizers}.
Although language-specific tokenizers can address these disparities,\footnote{%
Language-specific tokenizers have been shown to improve segmentation quality
and, after model adaptation, downstream performance for their target languages
\citep{rust-etal-2021-how, limisiewicz-etal-2023-tokenization}.}
using one with an arbitrary pretrained model requires either retraining the model
with a new tokenizer or adapting the model's parameters to a new vocabulary
\citep{downey-etal-2023-embedding,Remy_2024,feher-etal-2025-retrofitting}. 
Both approaches are computationally expensive, partially offsetting the benefit of starting from a pretrained model in the first place.

Importantly, the poor segmentations induced by English-centric tokenizers are not always due to missing vocabulary coverage: in many cases, the tokenizer already contains tokens that would support a more appropriate segmentation for the target language, but their parameters induce an alternative, suboptimal decomposition.\footnote{As an example, the German compound \texttt{Rathaus} (``city hall'') has the morphological segmentation \texttt{Rat} + \texttt{haus}. In a BPE tokenizer fit mostly on English text the merge rule \texttt{t} + \texttt{h} $\rightarrow$ \texttt{th} has a high rank and is applied first, so even if \texttt{Rat} and \texttt{haus} are both in the vocabulary, the segmentation is not possible.\looseness=-1}

In this paper, we propose Language-adaptive MAP (LangMAP) Tokenization, an inherently multilingual tokenization scheme. 
LangMAP follows the \unilm  tokenization algorithm \citep{kudo-2018-subword} %
in casting tokenization as recovering the latent segmentation of a string under a unigram model.
Given a dataset, \unilm learns a generative-model for it, composed of a vocabulary and its associated probability distribution. 
Uncovering a string's latent segmentation then consists of selecting the maximum a posteriori (MAP) segmentation under this model. 
LangMAP retains this formulation but replaces the single global distribution with a collection of language-specific distributions learned over the same shared vocabulary. At inference time, we compute the MAP segmentation under each language-specific distribution and select the one attaining the highest overall likelihood across all languages. 
A notable consequence is that LangMAP can adapt a pretrained model's tokenizer to individual languages without changing its vocabulary. Furthermore, although language labels are required to estimate the per-language distributions during estimation of LangMAP parameters, none are needed at tokenizer inference. %

We apply LangMAP to create language-specific variants of several open-source tokenizers. Across $7$ natural languages and $9$ programming languages, morphological boundary recall rises consistently for the four Latin-script natural languages. Alignment with AST leaf boundaries \cite{meister_tokenizer_analysis} rises for all programming languages on every tokenizer.
For the three non-Latin-script languages, the change in recall that LangMAP can induce is bottlenecked by how many entries from that script the vocabulary contains. 
We fine-tune Qwen2.5-0.5B and Llama-3.2-1B on five languages and measure grammatical acceptability with MultiBLiMP. 
For only one of the nine MultiBLiMP accuracy differences between LangMAP and the original tokenizer does the $95\%$ confidence interval exclude zero after multiple-testing corrections. 
This is consistent with prior findings that tokenizers whose segmentations match a language's morphology more closely do not necessarily give better downstream accuracy \cite{arnett2025alignment}. 
Three applications of LangMAP remain to be tested:\clara{wording} use for fine-tuning models in the coding domain, fine-tuning for languages whose pretraining data is limited, and training a model from scratch with LangMAP rather than replacing the tokenizer of an existing model.

\newcommand{\parameterspace}{\Phi}
\newcommand{\tokenisationclass}{\mathcal{H}}

\section{Tokenization}

Let $\str=\langle\stringunit_{1},\stringunit_2,\smalldots, \stringunit_N\rangle$ be a \defn{string}, which consists of characters (or bytes) from a base alphabet $\stralphabet$. 
Further, let $\vocab$ be a \defn{vocabulary}, a finite set composed of \defn{tokens} $\token\in\vocab$.\footnote{Tokens $\token$ are also sometimes called subwords; we avoid this naming because $\token$ need not align with orthographic words, in their typical definition. Further, a sequence of tokens $\tokens$ is also a string in the strict sense. We reserve the word \emph{string} for $\str$ throughout.} 
Each of these tokens $\token$ consists of a sequence of elements from $\stralphabet\cup\resalphabet$, where $\resalphabet$ denotes a finite set of reserved symbols (e.g., whitespace markers, punctuation, end-of-string markers, etc). Thus,
intuitively, each $\token\in\vocab$ is itself a finite string over the extended alphabet $\stralphabet\cup\resalphabet$.\looseness=-1

Tokenization is the process of converting a string $\str$ into a sequence of tokens $\tokens=\langle \token_{1},\smalldots,\token_{\toklength}\rangle$. 
This process, which we denote by the function $\tokfunc_{\unigramdistparams}\colon \stralphabet^* \rightarrow \vocab^*$,\footnote{$\{\}^*$ denotes the Kleene closure  of a set. In words, $\stralphabet^*$ is the set of all finite strings over the alphabet $\stralphabet$. } can be intuitively understood as creating a different representation of the original string. 
A tokenization algorithm typically defines a parametrized class of such maps $\tokenisationclass \defeq \{\tokfunc_{\unigramdistparams} \mid \unigramdistparams \in \parameterspace\}$, as well as a method for learning the parameters $\unigramdistparams \in \parameterspace$; these parameters then define (directly or indirectly) the tokenizer's vocabulary $\vocab$.
The application of $\tokfunc_{\unigramdistparams}$ to a string (once its parameters are learned) is referred to as \defn{inference}.

The map $\tokfunc_{\unigramdistparams}$ is typically designed to be lossless. 
Thus, given a sequence of tokens $\tokens = \tokfunc_{\unigramdistparams}(\str)$, we can reconstruct the input string $\str$ using a \defn{detokenization function} $\detokfunc\colon \vocab^* \rightarrow \stralphabet^*$.
Typically, $\detokfunc$ is as simple as concatenating the elements composing each of $\tokens$'s tokens, while handling its reserved symbols; formally, $\detokfunc(\tokens) = \token_{1}\circ\dots \circ \token_{\toklength}$, where $\circ$ denotes string concatenation and, when applied to tokens, it indicates that the strings forming each token are concatenated together.
Notably, $\detokfunc$ is generally non-injective: even when considering a fixed vocabulary $\vocab$, there are often multiple sequences $\tokens$ that yield the same $\str$ after the application of $\detokfunc$. We refer to all such token sequences as \defn{segmentations} of $\str$, and use $\allsegmentations_\vocab(\str) \defeq \{\tokens\in\vocab^*\colon \detokfunc(\tokens) = \str\}$ to refer to this set.
For a given $\tokfunc$, we refer to $\tokfunc(\str)$ as the \defn{canonical segmentation}
of $\str$ and to any other segmentation
$\tokens \in \allsegmentations_\vocab(\str)$ for which $\tokens \neq \tokfunc(\str)$ as a \defn{non-canonical segmentation}. 
That several token sequences detokenize to the same string has motivated researchers to propose alternative tokenizer inference schemes, which replace only $\tokfunc$---the tokenizer's inference rules---and keep the learned vocabulary unchanged. For example, the inference strategy may find the token sequence $\tokens\in \allsegmentations_\vocab(\str)$ of minimal length for a given input $\str$ \citep{schmidt-etal-2024-tokenization}, or may greedily select the longest in-vocabulary tokens \citep{hofmann-etal-2022-embarrassingly}. 
Our proposed method takes this approach, adapted for the multilingual setting.\looseness=-1

\subsection{\texorpdfstring{\unilm}{UnigramLM}}\label{sec:unigramlm}

The \unilm tokenization algorithm \citep{kudo-2018-subword} is one of the main tokenization algorithms, used, e.g., by the language models T5, PaLM 2, and XLNet \citep{t5,Anil2023PaLM2T,xlnet}. It frames tokenization as a probabilistic learning problem.
It assumes strings are generated via a two-stage generative model: latent token-sequences are sampled from a unigram distribution; these tokens are then concatenated to form strings. 
Tokenization then simply consists of uncovering a latent segmentation under this model (typically via a MAP approach).
As this generative model's parameters are unknown, \unilm also provides a method for estimating them. In the remainder of this section, we briefly describe the algorithm and refer to \cref{app:uni_background} for more details.

\paragraph{Generative Model.}
Let $\vocab$ be a vocabulary, and $\unigramdistparams \in \Delta^{|\vocab|-1}$ denote a probability distribution over it.
\unilm assumes strings are produced via a two stage generative model $P$.
First, token sequences $\tokens = \langle \token_1, \token_2, \smalldots, \token_{\toklength}\rangle$ are generated by sampling tokens independently from $\unigramdistparams$:%
\footnote{In order to be a valid probability distribution over token sequences, $P$ would need to include an explicit termination mechanism (e.g., an \texttt{EOS} symbol). Since this is omitted here, we use $\propto$ in \cref{eq:prob_token_sequence} to indicate that the expression is defined up to a normalization constant.}
\begin{equation} \label{eq:prob_token_sequence}
    \ptokenseq{\tokens} \defpropto \prod_{t=1}^{|\tokens|} \unigramdistparams[\token_t].
\end{equation}
where $\unigramdistparams[\token]$ denotes the probability of token $\token$ under $\unigramdistparams$.
Second, these sequences deterministically induce an observed string via a fixed detokenization function $\detokfunc$:
\begin{equation}
    P(\stringrv{=}\str \mid \tokenseqrv{=}\tokens) = 
    \left\{\begin{array}{lr}
    	1 & \texttt{if } \str = \detokfunc(\tokens) \\
		0 & \texttt{else}
    \end{array}\right.
\end{equation}

\paragraph{Inference Function.}
Given the parameters $\unigramdistparams$ of such a generative model, \unilm's inference procedure simply recovers the MAP 
segmentation of a string $\str$:
\begin{subequations}\label{eq:map_inference}
\begin{align}
    \bestsegmentation &\defeq \argmax_{\tokens \in \vocab^*} \post{\tokens}{\str}\\
    &=\argmax_{\tokens \in \allsegmentations_{\vocab}(\str)} 
    \prod_{t=1}^{|\tokens|} \unigramdistparams[\token_t]
\end{align}
\end{subequations}
(A detailed derivation is shown in \cref{sec:inference-derivation}.)
The solution to \cref{eq:map_inference} can then be found efficiently using a Viterbi-style dynamic program, which runs in $\mathcal{O}(N\cdot K_{\mathrm{max}})$ time, where $N$ is the length of the input string and $K_{\mathrm{max}}$ is the maximal length (in characters or bytes) of any token $\token\in\vocab$.

\newcommand{\dataset}{\mathcal{D}}

\paragraph{Parameter Estimation.}

Finally, \unilm also provides a way to learn the parameters $\unigramdistparams$ of its generative model.
Ideally, we would choose the $\unigramdistparams$ that maximize the log-likelihood of a target dataset $\corpus = \{\str_m\}_{m=1}^{M}$ under our generative model: i.e., $\likelihood(\corpus; \unigramdistparams) = \sum_{m=1}^M \log P(\stringrv{=}\str_m; \unigramdistparams)$.
As this is, in general, intractable, \unilm approximates $\unigramdistparams$ via the EM algorithm, which lower-bounds $\likelihood$ and maximizes that bound (\cref{sec:em-motivation}). 
Concretely, two steps are performed iteratively, either for a fixed number of steps or until changes to the objective value fall below some threshold.\footnote{$\unigramdistparams^{(0)}$ can be initialized as e.g., the uniform distribution.}
    First, under a fixed $\unigramdistparamscur$, the \textbf{E-Step}
  computes the expected count of each token in our corpus, i.e., the number of times it is used on average across all corpus segmentations:
  \newcommand{\hatc}[2]{\widehat{c}_{#1}(#2)} 
\begin{align}
   \hatc{\token}{&\corpus;\unigramdistparamscur} \defeq \\
   &\sum_{m=1}^M \sum_{\tokens\in\allsegmentations_{\vocab}(\str_m)}
    \postcur{\tokens}{\str_m}\,\countv{\token}{\tokens}\nonumber 
\end{align}
where $\countv{\token}{\tokens} \defeq \sum_{\tokindex=1}^{|\tokens|}\mathbbm{1}\{\token_\tokindex = \token\}$ counts the occurrences of token $\token$ in $\tokens$.
Second, the
\textbf{M-Step} sets each parameter to that token's normalized expected count:\looseness=-1
\newcommand{\unigramdistnewtoken}{\unigramdistparams^{(n+1)}[\token]}
\begin{align}\label{eq:mstep}
\unigramdistnewtoken
\;=\;\frac{\hatc{\token}{\corpus;\unigramdistparamscur}}
{\sum_{\token'\in{\vocab}}\hatc{\token'}{\corpus;\unigramdistparamscur}}.
\end{align}

When building a tokenizer, $\vocab$ is not known in advance, and \unilm selects it by initializing an over-sized set and pruning after each EM iteration (\cref{sec:vocab-pruning}). Our algorithm uses \unilm's parameter estimation but not its vocabulary selection: the vocabulary may equally be one learned by another algorithm, e.g., BPE.

\subsection{Tokenization in NLP Pipelines}

\unilm's estimation optimizes a tokenizer for the empirical distribution of its training corpus, so a corpus dominated by a few languages or domains yields a tokenizer implicitly optimized for those regimes. Such tokenizers will logically exhibit the cross-lingual disparities described in \cref{sec:intro}. 
These disparities are relevant, as over-segmentation leads to increased API costs  \citep{petrov-etal-2023-tokenizers}, and tokenizer segmentation quality has been shown to affect downstream generalization in multilingual models \citep[][\emph{inter alia}]{hofmann-etal-2021-superbizarre,klein-tsarfaty-2020-getting,liang-etal-2023-xlm,vemula-etal-2025-rethinking,arnett-bergen-2025-language,shani2026roots}.

A growing body of work has sought to mitigate these effects by modifying the tokenization stage or adapting pretrained models to alternative vocabularies. 
Broadly, existing approaches fall into two categories. 
The first class of approaches proposes new tokenization schemes.  
For example, \citet{foroutan-meister-et-al-2025-parity-aware-bpe} and \citet{limisiewicz-etal-2023-tokenization} propose variants of BPE and \unilm, respectively, that are particularly adapted for the multilingual setting. 
Some other works attempt to remove or relax the dependency on a fixed token vocabulary altogether. \tiago{should it be three classes of approaches, with this being the second? or does this fall into "new tokenization approaches"? maybe changing th wording of how this first class is described would help already}
Token-free and byte-level models operate directly on character or byte sequences \citep{xue-etal-2022-byt5, ahia2024magnet}, eliminating biases incurred  by vocabularies optimized for high-resource domains, but at the cost of substantially longer input sequences and increased training and inference complexity. 
The second class of approaches proposes post-training modifications to models. 
Post-hoc vocabulary adaptation methods, such as merging or retrofitting token embeddings for new vocabularies \citep{downey-etal-2023-embedding, Remy_2024,feher-etal-2025-retrofitting}, avoid full model retraining but still require either fine-tuning pretrained models or learning additional model parameters such as new token embeddings or auxiliary networks.

The next section introduces LangMAP, which gives every language its own parameters over one shared vocabulary. It is a tokenization scheme and an adaptation method at once: adapting a model to a further language changes neither the vocabulary nor the model's architecture.

\section{LangMAP Tokenization}\label{sec:langmap}

We propose Language‑adaptive MAP (LangMAP) Tokenization: a new tokenization algorithm that produces segmentations optimized for each language in its training set.\footnote{We note that the algorithm is generally applicable to any partition of the corpus; we focus on languages, as LM performance according to this partition is of particular concern to the community, but perform evaluations on code as well.} Notably, LangMAP requires language labels during estimation, but not at inference. LangMAP can be used either when training a language model from scratch or to adapt an already pretrained model, accepting any base vocabulary (e.g., one learned by BPE, or that of a pretrained model) without modifying it.

\subsection{Learning Language‑Specific Parameters}\label{sec:lasulm}
Let $\languages$ be the set of languages under consideration and $\lang\in \languages$ be a single language. 
Let $\corpus^\lang$ denote the subset of the corpus corresponding to language $\lang$. 
Further, let $\vocab$ be a fixed base vocabulary, e.g., the vocabulary of a pretrained language model or one learned by a standard tokenization algorithm such as \unilm.
Keeping this vocabulary fixed, we learn language-specific unigram parameters\footnote{We re-initialize $\unigramdistparams^{(0)}_\lang$ as the current token frequency distribution in the corpus, as estimated by \unilm's EM procedure.} $\unigramdistparamslang$ by applying the EM procedure described in \cref{sec:unigramlm} independently to each language-specific corpus $\corpus^\lang$. 
For each language, the expected token counts in the E-step are computed (exclusively) over strings in $\corpus^\lang$, 
while the M-step normalizes these counts over the full (shared)  vocabulary $\vocab$. 
As mentioned above, as $\vocab$ is fixed, there is no vocabulary learning, and the pruning step of the standard \unilm algorithm is skipped; only token probabilities are estimated.

\subsection{Language-Adaptive Inference}
At inference time, we find the most probable segmentation for each language $\lang \in \languages$ under the unigram model with parameters $\unigramdistparamslang$.
We then compare segmentations across all languages, choosing the one with maximum unigram likelihood according to any $\unigramdistparamslang$.
Notably, this second step implicitly identifies the string's language using UniLID, a recent method introduced by \citet{meister2026languagethisasktokenizer},\footnote{Explicitly, UniLID performs language identification using an approach identical to \cref{eq:langmap_inference}, but returning $\lang$ instead of $\tokens_\lang$.} and then returns its segmentation.
Formally:
\begin{align}
    &\!\!\!\underbrace{\tokens_\lang = \argmax_{\tokens \in \allsegmentations_{\vocab}(\str)} \,\, \ptokenseqlang{\tokens}}_{\text{Standard per-language inference via UnigramLM}}\\
&\underbrace{\tokens^* = \argmax_{\tokens_\lang \,|\, \lang\in \languages} \,\,\ptokenseqlang{\tokens_\lang}}_{\text{Implicit language-identification via UniLID}} \label{eq:langmap_inference}
\end{align}
Although this formulation appears to require performing the \unilm inference problem for each language, inference remains efficient due to the shared vocabulary. 
Because $\vocab$ is fixed, the segmentation lattice, which provides all valid segmentations of the input string under $\vocab$, is identical across languages and needs to be constructed only once.
The per-language instantiations of the problem differ only in the edge weights induced by the language-specific token probabilities.
As a result, the overall complexity scales linearly with the number of languages, requiring an additional $\mathcal{O}(|\languages|)$ passes over a fixed lattice, rather than reconstructing the lattice itself for each model. 
This overhead could be reduced, if desired, by using coarser-grained categorizations, e.g., language families instead of individual languages.

\subsection{Computational Complexity, Scope, and Limitations}

From a computational standpoint, LangMAP differs from the standard \unilm tokenization algorithm only in that it maintains a separate distribution per language rather than a single shared one; its cost follows directly from this. 
Training amounts to running the same EM procedure once per language, each over only that language's portion of the corpus, so the total work is comparable to a single \unilm run; the $|\languages|$ fits are moreover independent and completely parallelizable.
Empirically, we observed that the estimates stabilized after about one thousand documents per language, well below the $25$M tokens per language on which every distribution in this paper was fit (\cref{ap:convergence}).

At inference, the segmentation lattice depends only on the (fixed) vocabulary and the input string, so it is built once and reused across all language-conditional distributions; LangMAP therefore adds only an $\mathcal{O}(|\languages|)$ factor over standard \unilm decoding: one Viterbi-style pass per language over the shared lattice, plus an $\mathcal{O}(|\languages|)$ comparison to select the most likely segmentation.
Finally, because each distribution is fit independently and the vocabulary is held fixed, incremental extension to new languages or dialects can be done easily and efficiently; only the unigram distribution of the new language needs to be learned.\looseness=-1

\section{Experiments}

We evaluate whether LangMAP segmentations improve tokenization \emph{intrinsically} (measured as, e.g., compression) and \emph{extrinsically} (measured as downstream performance after model fine-tuning).
For each base tokenizer we compare intrinsic and downstream metrics under the unchanged base tokenizer
(\textsc{Original}) and \textsc{LangMAP}.

\subsection{Tokenizers, languages and data}\label{setup}
\paragraph{Tokenizers and models.}
\cref{tab:tokenizers} in \cref{app:exp_setup} lists the base models whose tokenizers we study here (14 rows, covering 16 models: the Llama 3 row covers Llama-3.1-8B and Llama-3.2-1B, and the Qwen row covers Qwen2.5 and Qwen3). 
This list spans models from the Gemma, Llama, Mistral, and Qwen families (among others), covering
distinct tokenization algorithms and vocabulary sizes.
Notably, several of the model families we study use identical tokenizers across different sizes and/or versions (e.g.\ Gemma~2 with 2b or 9b parameters use the same tokenizer, or
Llama-3.1 with 8b and Llama-3.2 with 1b).
We therefore report \emph{deduplicated} intrinsic metrics, over distinct tokenizers rather than over models. \Cref{tab:morphscore} reports MorphScore per (model family, language), measured on the LangMAP tokenizers built over each model's own vocabulary that the downstream experiments of \cref{sec:extrinsic-results} train with, so that the intrinsic and downstream measurements describe the same tokenizers. The wider deduplicated sweep, $10$ tokenizers on each of $7$ languages, is reported in \cref{tab:app-sweep-recall}.\looseness=-1

\paragraph{Languages and training data.}
We measure morphological alignment on seven natural languages, each of which has the Universal Dependencies treebank that the metric requires: German, Finnish, Hungarian, Turkish, Hindi, Arabic and Tamil. Bengali and Thai appear only in MorphScore evaluations and the label-free routing measurement of \cref{tab:routing}, and Swahili, which has no treebank, only in the additional intrinsic metrics of \cref{app:add_nl_intrinsic}. \Cref{app:exp_setup} lists all ten with their script and morphological type.
We also evaluate tokenizers on nine programming languages: C, C++, C\#, Go, Java, JavaScript, PHP, Python, and TypeScript, chosen to span brace-delimited vs.\ whitespace-sensitive and statically vs.\ dynamically typed conventions.
The code measurements cover eight model families, listed in \cref{tab:code-overall-grid}.
For LangMAP parameter estimation, we use data from FineWeb2 \citep{penedo2025fineweb2} and The Stack \citep{lozhkov2024starcoder}.\footnote{\url{https://huggingface.co/datasets/bigcode/the-stack}}

\paragraph{Estimation.}
We learn each $\unigramdistparamslang$---the per-language unigram distribution over the original tokenizer's vocabulary---with the EM procedure of
\cref{sec:langmap}, run for $K{=}10$ iterations. For each language, we use $25M$ tokens from the aforementioned datasets (FineWeb2 for natural language; The Stack for code) to perform this parameter estimation. 
We initialize $\unigramdistparamslang$ from a unigram distribution estimated jointly over the entire corpus via the EM procedure detailed in \cref{sec:unigramlm}.

\subsection{Tokenizer configurations}\label{sec:configurations}
We compare five tokenizer configurations at the same training budget.
\textsc{Original} keeps the model's own tokenizer.
\textsc{LangMAP} re-scores that same vocabulary per language.
\textsc{Base-Unigram} applies \unilm segmentation over the same vocabulary with one set of language-agnostic scores, which separates the effect of the language-specific scores from the effect of switching to Unigram segmentation at all.
We construct a \textsc{Placebo} variant as an additional baseline: per-language distributions are set by sorting the vocabulary by LangMAP score, dividing the sorted list into buckets of $500$ pieces, and permuting the scores within each bucket. %
\textsc{Vocabulary-extension} is the baseline for the alternative to LangMAP: changing which pieces the vocabulary contains, rather than changing the scores over a fixed vocabulary. 
It broadly follows the approach first proposed by \citet{wang-etal-2020-extending} and \citet{chau-etal-2020-parsing}. 
Concretely, we fit a monolingual $32$k BPE tokenizer on the target language, drop the pieces already present in the base vocabulary and the pieces made of digits ($165$ to $210$ per language), append the $16{,}000$ most frequent of the remainder, and initialize each new embedding row from the mean of the embeddings of that piece's decomposition under the original tokenizer \citep[following ][]{dobler-de-melo-2023-focus}. 
This adds $14.1$M trainable parameters (see \cref{app:exp_setup}). The other four tokenizer configurations leave the parameter count unchanged, so the comparison between \textsc{LangMAP} and \textsc{Vocabulary-extension} is the comparison between re-scoring a vocabulary at no parameter cost and extending it at a parameter cost.
\Cref{tab:f15-multiblimp} reports Qwen2.5-0.5B under the different configurations. Llama-3.2-1B was run with \textsc{Original} and \textsc{LangMAP} only.

\subsection{Adaptation protocol}\label{sec:adaptation}
Extrinsic evaluations measure downstream model performance when using one of the above configurations. 
For these experiments, we fine-tune our base models on new language data, tokenized using the specified tokenizer.\footnote{Using LangMAP out-of-the-box on a language already in the model's pretraining data without any continued pre-training or fine-tuning of the model systematically leads to worse performance; most probability mass for a string is placed on the tokenization the model saw during training, i.e., the canonical tokenization under the base tokenizer \citep{cao-rimell-2021-evaluate}, and so this behavior is to be expected.}
Due to computational constraints, we restrict these experiments to a subset of languages and a subset of the models in \cref{tab:tokenizers}.
We include Hindi as a negative control. Before running the experiments, we found that neither vocabulary had sufficient coverage of Hindi pieces to allow LangMAP to meaningfully change the Hindi segmentation. The Qwen2.5-0.5B vocabulary holds 67 Devanagari pieces, so LangMAP segments Hindi at the character level; the Llama-3.2-1B vocabulary holds 958 Devanagari pieces, and LangMAP's Hindi segmentation differs from the original tokenizer's on few word types: the MorphScore recall difference is +0.0010 (Table 1). We therefore expected LangMAP to have little effect on Hindi in the downstream experiments. In contrast, we expected vocabulary extension to have a measurable effect. We also repeat the experiments on Llama-3.2-1B, using Turkish, Arabic, Hindi, and Hungarian, to see whether the results generalize to another model.
For each combination of base model, target language, and tokenizer, we fine-tune the model on target-language FineWeb2 text, with English text making up $20\%$ of the training data.
We fine-tune with LoRA of rank $16$, $\alpha{=}32$, dropout $0.05$, learning rate $2{\times}10^{-4}$, and fully-trainable tied embeddings.
Every run takes the same number of optimizer steps. 
\cref{app:exp_setup} gives more detail on how controls were implemented. 
We run 3 seeds per (language, tokenizer configuration).\footnote{We extend to 5 seeds for Hungarian under a rule specified before running the experiments: namely, seeds were added when a difference has a $95\%$ confidence interval excluding zero before correction.}
The Llama-3.2-1B runs use the same exact training settings and data; only the base model and its vocabulary differ.

\subsection{Intrinsic measurements}
We evaluate LangMAP (\textsc{LangMAP}) against the base tokenizer \textsc{Original} using the TokEval metric suite,\footnote{\url{https://github.com/cimeister/tokenizer-intrinsic-evals}} comparing paired (\textsc{Original}, \textsc{LangMAP}) tokenizations of the same input texts.
For natural languages, we evaluate on the FLORES-200 parallel dataset \citep{nllb2022flores}.
For code metrics, we report on StarCoder data \citep{li2023starcoder}.
We report three of these metrics here in the body:
\begin{itemize}[noitemsep,leftmargin=*,topsep=0pt]
    \item \defn{MorphScore} compares the boundaries a tokenizer places inside a word type with that word type's gold morpheme boundaries; we report its recall, the fraction of gold boundaries the tokenizer also places, together with its precision and micro-F1, over the Universal Dependencies subsets \citep{arnett2025alignment,nivre-etal-2017-universal}.
\item \defn{AST leaf-boundary alignment} is the fraction of parse-tree leaf spans whose start and end both coincide with a token boundary.
\item \defn{Identifier fragmentation} is the fraction of programmer-defined identifiers split into more than one token. 
\end{itemize}
R\'{e}nyi entropy, bigram entropy, fertility and compression rate are reported in \cref{app:int_metrics}. 
We refer the reader to the TokEval library for descriptions and implementation details.

For all metrics, we report paired differences: $\Delta=\textsc{LangMAP}-\textsc{Original}$; the   direction that indicates an improvement when using LangMAP is noted per caption. We also report 95\% confidence intervals throughout. 
Statistical significance is assessed via a paired percentile bootstrap, with 10,000
resamples over (tokenizer, language) pairs; see \cref{app:exp_setup} for details.
The per-(model family, language) MorphScore values of \cref{tab:morphscore} use a different resampling unit (see \cref{app:morphscore-details}).

\subsection{Downstream measurements}
We report three downstream measurements: 
\begin{itemize}[noitemsep,leftmargin=*,topsep=0pt]
\item Target-language \defn{MultiBLiMP accuracy} \citep{jumelet2026multiblimp}, evaluated 0-shot (i.e., no worked examples in the prompt)
\item \defn{English Belebele accuracy} \citep{bandarkar2024belebele} as a measure of English retention, evaluated 5-shot
\item \defn{FLORES+ bits per byte} in the target language and in English
\end{itemize}
MultiBLiMP is an automatically-constructed minimal-pair benchmark, derived from Universal Dependencies treebanks; each of its items presents two near-identical sentences differing only on a targeted morphosyntactic feature, and the model is scored on whether it assigns higher likelihood to the grammatical one. 
Belebele evaluates reading comprehension via four-way multiple-choice questions grounded in FLORES-200 passages.
Target-language Belebele and Global-PIQA \citep{chang2025global} are not reported because their accuracies in these languages are at or near chance at this model scale. 
We evaluate the pretrained model (\textsc{No-adaptation}), using its original tokenizer and without any fine-tuning. We evaluate it on the same 0-shot MultiBLiMP and 5-shot English Belebele tasks. This baseline shows how much performance changes after adapting the model to a new tokenizer. For each MultiBLiMP difference from the original tokenizer, we report the mean over seeds, the standard deviation across seeds, and a paired percentile bootstrap over items with seeds as replicates ($10{,}000$ resamples). Multiple testing is controlled for by the Holm procedure at $\alpha=0.05$: over the $18$ differences from the original tokenizer in Qwen2.5-0.5B, and over the $4$ in Llama-3.2-1B. The bits-per-byte and Belebele numbers are reported as means over seeds. \Cref{app:exp_setup} states further evaluation details.

\begin{table}[t]
\centering\small\setlength{\tabcolsep}{4pt}
\adjustbox{max width=\columnwidth}{%
\begin{tabular}{@{}lrccc@{}}
\toprule
\textbf{Lang.} & $n$ & $\Delta$\textbf{micro-F1} & $\Delta$\textbf{Rec.} & $\Delta$\textbf{Prec.} \\
\midrule
\multicolumn{5}{@{}l}{\emph{Qwen2.5-0.5B}} \\
Turkish              & 32{,}560 & $+0.0688^{\star}$ & $+0.2108^{\star}$ & $+0.0703^{\star}$ \\
Finnish              & 40{,}776 & $+0.0245^{\star}$ & $+0.1332^{\star}$ & $+0.0391^{\star}$ \\
Hungarian            &  7{,}924 & $+0.0350^{\star}$ & $+0.1530^{\star}$ & $+0.0385^{\star}$ \\
Arabic               & 11{,}256 & $+0.0743^{\star}$ & $+0.0663^{\star}$ & $+0.0350^{\star}$ \\
Hindi$^{\dagger}$    &  4{,}723 & $-0.0116^{\star}$ & $+0.1008^{\star}$ & $+0.0217^{\star}$ \\
\midrule
\multicolumn{5}{@{}l}{\emph{Llama-3.2-1B}} \\
Turkish              & 32{,}560 & $+0.0884^{\star}$ & $+0.1937^{\star}$ & $+0.0743^{\star}$ \\
Finnish              & 40{,}776 & $+0.0225^{\star}$ & $+0.1151^{\star}$ & $+0.0319^{\star}$ \\
Hungarian            &  7{,}924 & $+0.0413^{\star}$ & $+0.1659^{\star}$ & $+0.0440^{\star}$ \\
Arabic               & 10{,}603 & $+0.0074^{\star}$ & $-0.0043$         & $+0.0015$ \\
Hindi                &  4{,}723 & $-0.0009^{\star}$ & $+0.0010^{\star}$ & $-0.0000$ \\
\bottomrule
\end{tabular}}
\caption{MorphScore difference (\textsc{LangMAP} minus \textsc{Original}) on the tokenizers the models of \cref{sec:extrinsic-results} were fine-tuned with; higher is better in all three columns.  $^{\star}$ marks a $95\%$ confidence interval excluding zero. $^{\dagger}$ Qwen2.5-0.5B Hindi recall is $1.0000$ by construction: that vocabulary holds only $67$ Devanagari pieces. $n$ is the number of word types scored, which is also the sample size of the paired bootstrap. \Cref{app:morphscore-details} gives further details, including significance testing protocol and the macro-F1 column.}
\label{tab:morphscore}
\vspace{-5pt}
\end{table}

\section{Results}

\subsection{Morphological Alignment}
\Cref{tab:morphscore} reports MorphScore measured on the LangMAP tokenizers that the fine-tuned models of \cref{sec:extrinsic-results} were trained with. 
Recall rises in nine of the ten recall measurements, and the $95\%$ confidence interval excludes zero in each of those nine. However, MorphScore recall on Arabic with the Llama-3.2-1B tokenizer is negatively impacted with an interval spanning zero. The increase in recall means that LangMAP places more token boundaries near gold morpheme boundaries, while the simultaneous increase in precision suggests that these changes bring the segmentation closer to morphologically meaningful units rather than simply adding noise.
 However, we have reason to believe that not all observed $\Delta$ values reflect changes caused by the tokenizer itself. Namely, the two Arabic results are potentially properties of the MorphScore gold morpheme references (see \cref{app:morphscore-details}) 
and the Qwen2.5-0.5B Hindi result is largely determined by the tokenizer's limited Devanagari coverage: its vocabulary contains only $67$ Devanagari pieces. With so few pieces available, any tokenizer restricted to this vocabulary must segment Hindi largely at the character level, giving a recall of $1.0000$ by construction. The Llama-3.2 vocabulary contains more Devanagari pieces ($958$), but the Hindi $\Delta$ values are still $0$. This is likely because $958$ pieces remain too few to allow substantially different segmentations. In this case, LangMAP makes few or no changes to the original segmentation, so a $\Delta$ of $0$ simply indicates that the two segmentations are unchanged. \Cref{app:add_nl_intrinsic} provides a fuller analysis. These results also illustrate a limitation of LangMAP. As LangMAP re-scores a fixed vocabulary rather than adding new tokens, it can only change how the existing pieces are combined. The extent of this change therefore depends on how well the original vocabulary covers the language in the first place.

A wider sweep is reported in \cref{tab:app-sweep-recall}: $10$ distinct tokenizers on $7$ languages, adding German and Tamil to the five above. For all $40$ recall measurements on the four Latin-script languages, the recall difference is positive. For the three non-Latin-script languages, diagnostics are performed. Results for the other intrinsic metrics are in \cref{app:add_nl_intrinsic}.

\crnew{%
\begin{table}[t]
\centering\small
\adjustbox{max width=\columnwidth}{%
\begin{tabular}{lrrr}
\toprule
Metric & $n$ & Improved & $\Delta$ \\
\midrule
Overall AST alignment   & 72 & $100\%$ & $+0.027^{\star}$ \\
Identifier alignment    & 72 & $99\%$  & $+0.028^{\star}$ \\
Operator alignment      & 72 & $97\%$  & $+0.025^{\star}$ \\
Delimiter alignment     & 72 & $100\%$ & $+0.030^{\star}$ \\
Literal alignment       & 72 & $93\%$  & $+0.029^{\star}$ \\
Keyword alignment       & 72 & $51\%$  & $+0.003^{\star}$ \\
\midrule
Ident.\ frag.\ rate     & 72 & $94\%$  & $-0.009^{\star}$ \\
Mean tokens / ident.    & 72 & $86\%$  & $-0.024^{\star}$ \\
\bottomrule
\end{tabular}}
\caption{Change in intrinsic tokenizer code metrics (\textsc{LangMAP} minus \textsc{Original}) over $8$ model families and
$9$ programming languages. The alignment rows are AST alignment stratified by node category; higher is better. \emph{Mean tokens / ident.}
is the mean number of tokens per identifier; lower is better for both this metric and \emph{Ident. frag. rate}. \emph{Improved} is the percentage
of the $n$ measurements in which the difference improves; $^{\star}$ marks a $95\%$
confidence interval excluding zero. }
\label{tab:code-summary}
\vspace{-10pt}
\end{table}}

\subsection{Code Tokenization}
\label{sec:code-results}
\Cref{tab:code-summary} reports the paired differences between LangMAP and the original tokenizer, averaged across $8$ open-source tokenizers and $9$ programming languages. LangMAP consistently improves alignment with AST leaf boundaries: the change is positive for every tokenizer-language pair. LangMAP also reduces the fraction of programmer-defined identifiers that are split across multiple tokens by $0.009$, and reduces identifier fertility (Mean tokens / ident). \Cref{app:add_code_intrinsic} reports the evaluation set, bootstrapping procedure used to compute confidence intervals, and absolute alignment scores.

\subsection{Extrinsic Results}
\label{sec:extrinsic-results} %

\crnew{%
\begin{table*}[t]
\centering\small
\setlength{\tabcolsep}{3pt}
\begin{tabular}{@{}l c ccccc@{}}
\toprule
Language & No-adapt. & Original & Base-Unigram & LangMAP & Placebo & Vocab.-ext. \\
\midrule
\multicolumn{7}{@{}l}{\emph{Qwen2.5-0.5B}} \\
Finnish   & 0.6420 & 0.8734 {\scriptsize$\pm 0.0062$} & 0.8694 {\scriptsize$\pm 0.0020$} & 0.8712 {\scriptsize$\pm 0.0039$} & -- & $0.9128^{\star}$ {\scriptsize$\pm 0.0010$} \\
Hungarian & 0.7763 & $0.9527^{\dagger}$ {\scriptsize$\pm 0.0036$} & 0.9538 {\scriptsize$\pm 0.0083$} & $0.9643^{\star\dagger}$ {\scriptsize$\pm 0.0023$} & $0.9564^{\dagger}$ {\scriptsize$\pm 0.0054$} & $0.9612^{\dagger}$ {\scriptsize$\pm 0.0039$} \\
Turkish   & 0.7560 & 0.8387 {\scriptsize$\pm 0.0035$} & 0.8305 {\scriptsize$\pm 0.0029$} & 0.8308 {\scriptsize$\pm 0.0108$} & -- & $0.8603^{\star}$ {\scriptsize$\pm 0.0058$} \\
Arabic    & 0.9185 & 0.9248 {\scriptsize$\pm 0.0063$} & 0.9300 {\scriptsize$\pm 0.0033$} & 0.9320 {\scriptsize$\pm 0.0026$} & 0.9213 {\scriptsize$\pm 0.0050$} & 0.9262 {\scriptsize$\pm 0.0048$} \\
Hindi     & 0.8493 & 0.9781 {\scriptsize$\pm 0.0014$} & 0.9777 {\scriptsize$\pm 0.0051$} & 0.9790 {\scriptsize$\pm 0.0029$} & 0.9790 {\scriptsize$\pm 0.0022$} & 0.9751 {\scriptsize$\pm 0.0007$} \\
\midrule
\multicolumn{7}{@{}l}{\emph{Llama-3.2-1B}} \\
Turkish   & 0.8416 & 0.8693 {\scriptsize$\pm 0.0063$} & -- & 0.8588 {\scriptsize$\pm 0.0041$} & -- & -- \\
Hungarian & 0.9325 & $0.9382^{\dagger}$ {\scriptsize$\pm 0.0105$} & -- & $0.9463^{\dagger}$ {\scriptsize$\pm 0.0030$} & -- & -- \\
Arabic    & 0.8897 & 0.9073 {\scriptsize$\pm 0.0042$} & -- & 0.9078 {\scriptsize$\pm 0.0029$} & -- & -- \\
Hindi     & 0.9806 & 0.9719 {\scriptsize$\pm 0.0052$} & -- & 0.9717 {\scriptsize$\pm 0.0038$} & -- & -- \\
\bottomrule
\end{tabular}
\caption{Target-language MultiBLiMP accuracy after adapting each model to the given language at a matched training budget; higher is better. \emph{No-adapt.} is the pretrained model with its own tokenizer and no fine-tuning (one evaluation, no seeds). Every other value is the mean over 3 seeds, with the $\pm$ value giving the standard deviation across seeds; $\dagger$ marks a value computed over 5 seeds (all Hungarian values except Qwen2.5-0.5B Base-Unigram). $\star$ marks a value whose difference from Original has a 95\% confidence interval excluding zero after Holm correction. \looseness=-1}
\label{tab:f15-multiblimp}
\end{table*}}

\paragraph{Adapting a language model.}
\Cref{tab:f15-multiblimp} reports target-language MultiBLiMP accuracy after adapting Qwen2.5-0.5B and Llama-3.2-1B to a single language at a time; multiple testing corrections are performed in all significance tests.
Notably, the improvement from fine-tuning alone is larger than any improvements that come from the choice of tokenizer. 
\textsc{Vocabulary-extension} is the only configuration whose accuracy is significantly above the original tokenizer's in more than one language (in Finnish and Turkish). 
Hungarian under the Qwen2.5-0.5B base model is the one setting where LangMAP offers a significant MultiBLiMP improvement over the model's original tokenizer, i.e.,  its $95\%$ confidence interval excludes zero after correction.
Two control configurations were run in Hungarian.
The model using the \textsc{Base-Unigram} tokenizer scores $+0.001$ accuracy above the original tokenizer in Hungarian, so the improvement in Hungarian MultiBLiMP is not produced by the switch to \unilm segmentation.
The model using the \textsc{Placebo} tokenizer, the score permutation baseline described in \cref{sec:configurations}, is $+0.004$ above the model that uses the original tokenizer. The \textsc{LangMAP}-minus-\textsc{Placebo} difference is $+0.008$; its $95\%$ confidence interval excludes zero before correction for multiple comparisons but not after it.
The Hungarian result is not reproduced in the second model family: no Llama-3.2-1B difference has an interval excluding zero after correction. 
\cref{app:downstream} contains an analysis looking at whether per-language MorphScore differences are positively associated with downstream MultiBLiMP differences. 
There is no significant correlation between the two in our results: over the nine adapted (language, model family) combinations, the Spearman correlation is negative under all four MorphScore columns, and none of the correlations is distinguishable from zero at $n=9$.

\paragraph{Bits per byte and English retention.}
\Cref{tab:f15-bpb} reports FLORES+ bits per byte and pooled English Belebele accuracy for Qwen2.5-0.5B, and \cref{tab:f15-llama-bpb} repeats the bits-per-byte measurement on Llama-3.2-1B.
Only \textsc{Vocabulary-extension} improves target-language likelihood: models trained with it reach the lowest target-language bits per byte in every language, at a modest cost in English bits per byte.
Where the vocabulary is left unchanged and only segmentations are re-scored, likelihood gets worse rather than better: models trained with the \textsc{LangMAP} and \textsc{Base-Unigram} tokenizers are above the original tokenizer's bits per byte in every target language and in English.
We conclude that adaptation with LangMAP does not improve model likelihood at this scale and budget.
In terms of retention of English performance, Belebele accuracies differ more between the adapted and the unadapted model than between tokenizer configurations. Qwen2.5-0.5B without adaptation scores $0.573$, against $0.481$ to $0.500$ for the five adapted configurations; the drop of $0.07$ to $0.09$ accuracy is shared across all five configurations. The spread between the five is within the range of their pooled standard deviations, though, implying there are not significant differences between these approaches' impacts on English at this scale.  

\paragraph{Adapting to five languages jointly.}
Adapting one model per language is not the setting LangMAP is designed for: a shared vocabulary with one distribution per language is intended for one model serving several languages.
\Cref{tab:f15-joint} describes this protocol, in which one model is adapted to all five languages at once (equal mix, plus English at proportion $0.2$) and the segmentation is selected at inference without a language label.
We observe empirically that label-free routing mechanism, i.e., the selection of a string's segmentation without a language label (\cref{eq:langmap_inference}), corresponds perfectly to that string's true language. 
\Cref{tab:routing} reports an analysis of this label-free routing mechanism over $12$ languages and truncated inputs on a separate base model. Models trained with the LangMAP tokenizer are slightly above the original tokenizer's accuracy in three languages under Qwen2.5-0.5B and indistinguishable from it in the other two. These improvements do not repeat under Llama-3.2-1B. 
In short, while the per-language segmentation scheme works, our results about whether the scheme aids in downstream performance are inconclusive.

\section{Conclusion}
We introduced LangMAP, a multilingual tokenization scheme that extends \unilm by maintaining a separate unigram distribution per language over a single, fixed vocabulary.
LangMAP can be trained as a tokenizer in its own right or used to adapt a tokenizer of an existing pre-trained model without altering its vocabulary or parameters.
In either case, language labels are required only during estimation; at inference, the model selects the appropriate language-specific segmentation without knowing the input language.
On natural language, LangMAP raised the alignment of token boundaries with gold morpheme boundaries consistently for Latin-script languages. For programming languages, it raised alignment with abstract syntax tree leaf boundaries.
Limited fine-tuning results do not show consistent improvements from using LangMAP, leaving downstream applications of LangMAP as an open research question.

\section*{Limitations}
While the use of a fixed shared vocabulary ensures compatibility with pretrained models, it also bounds the representational capacity of the tokenizer.
If linguistically appropriate tokens for a given language are absent from the vocabulary, language-specific reweighting of token probabilities alone cannot recover good segmentations for those languages. This limitation is visible in our results.
A natural follow up would be to pair LangMAP with vocabulary extension so that adaptation is no longer confined to the tokens a model already supports.
Another limitation is that our method requires labeled training data to estimate language-specific distributions.
We observe empirically that very few labeled samples are needed to get good unigram distribution estimates, but the method does not work in the complete absence of labels.
We leave unsupervised or weakly supervised extensions for future work.
Finally, our downstream evidence is narrow. It covers two model families at one size each (Qwen2.5-0.5B on five languages, Llama-3.2-1B on four), one adaptation budget, and 3 seeds per (language, tokenizer configuration).  
Bits per byte is higher for LangMAP than for the original tokenizer in every language and both the models that we fine-tuned, so we make no claim about its ability to improve base language modeling capabilities.\looseness=-1

\section*{Acknowledgments}
We thank Kris Cao for helpful conversations at the beginning of this project.

\bibliography{bibliography}

%%%%%%%%%%%%%%%%%%%%%%%%%%%%%%%%%%%%%%%%%%%%%%%%%%%%%%%%%%%%%%%%%%%%%%%%%%%%%%%
%%%%%%%%%%%%%%%%%%%%%%%%%%%%%%%%%%%%%%%%%%%%%%%%%%%%%%%%%%%%%%%%%%%%%%%%%%%%%%%
% APPENDIX
%%%%%%%%%%%%%%%%%%%%%%%%%%%%%%%%%%%%%%%%%%%%%%%%%%%%%%%%%%%%%%%%%%%%%%%%%%%%%%%
%%%%%%%%%%%%%%%%%%%%%%%%%%%%%%%%%%%%%%%%%%%%%%%%%%%%%%%%%%%%%%%%%%%%%%%%%%%%%%%
\newpage
\clearpage
\appendix
\crefalias{section}{appendix}
\crefalias{subsection}{appendix}

\providecommand{\crnew}[1]{{\color{red}#1}}

\section{Experimental Setup Details}\label{app:exp_setup}

\paragraph{Evaluated models and their tokenizers.} 
\Cref{tab:tokenizers} contains information about our evaluated models and their tokenizers.

\begin{table}[h]
\centering
\small
\caption{Tokenizer and Vocabulary of Base Model tokenizers that are used for intrinsic and downstream evaluation. We additionally report results for \texttt{OLMo-3-7B} and
\texttt{EuroLLM-1.7B}. The Llama 3 and Qwen 2.5 \& 3 rows each cover two model versions that share one tokenizer. $|V|$ is the full vocabulary size, added and special tokens included.}
\label{tab:tokenizers}
\begin{tabular}{llr}
\toprule
\textbf{Model} & \textbf{Algorithm} & $|V|$  \\
\midrule
Bloom-3b            & BPE                   & 250,880  \\
Gemma-3-1b    & BPE     & 262,144  \\
Gemma-2-2b   & BPE  & 256,000  \\
Granite 3.0-8b      & BPE                   & 49,152   \\
Llama 3       & BPE     & 128,256 \\
Mistral-7b-v0.3     & BPE & 32,768   \\
Mistral-Nemo-12b    & BPE & 131,072 \\
Phi-3-mini               & BPE & 32,064   \\
Phi-4               & BPE & 100,352  \\
Qwen 2.5 \& 3         & BPE              & 151,936  \\
XGLM-2.9b          & UnigramLM  & 256,008 \\
Yi-1.5-6b           & BPE &64,000 \\
\midrule 
\multicolumn{3}{@{}l}{\textit{Appendix-only Models}} \\
OLMo-3-7B         & BPE       & 100{,}278 \\
EuroLLM-1.7B      & BPE       & 128{,}000 \\

\bottomrule
\end{tabular}
\end{table}

\newcommand{\langmap}{LangMAP}
\newcommand{\base}{Original}
\paragraph{Paired-bootstrap differences.} All reported $\Delta$ values are paired comparisons over (tokenizer, language)
pairs. For each pair, \base{} and \langmap{} are evaluated on the same test
data, and we take the per-pair difference
$\Delta_c = m(\langmap_c) - m(\base_c)$ for metric $m$. 
We report the aggregate effect as the mean over the $n$ pairs,
$\bar{\Delta} = \tfrac{1}{n}\sum_{c=1}^{n}\Delta_c$, and estimate 95\% confidence
intervals with a nonparametric percentile bootstrap that resamples the $n$ pairs
with replacement: we draw 10,000 resamples, recompute $\bar{\Delta}$ on each,
and report the 2.5th and 97.5th percentiles of the resulting distribution. 
An aggregate effect is significant when this interval excludes zero.\looseness=-1

\paragraph{Languages.}
The seven languages measured for morphological alignment are German (fusional, Latin
script), Finnish, Hungarian and Turkish (agglutinative, Latin script), Hindi (mixed,
Devanagari), Arabic (templatic, Arabic script) and Tamil (agglutinative, Tamil script). Each
has the Universal Dependencies treebank that the metric requires. Bengali (agglutinative,
Bengali script) and Thai (analytic, Thai script) have treebanks as well and are measured only
in the label-free routing of \cref{tab:routing}. Swahili has no treebank and appears in
\cref{tab:app-surface-perlang} only.

\paragraph{Training budget and seeds.}
The adaptation budget is defined in padded training positions rather than in real
tokens. Every sequence is padded to $1024$ positions, and every run takes the same $3{,}561$
optimizer steps at $49{,}152$ positions per step, $175{,}030{,}272$ positions in total. Fixing
the number of optimizer steps is what makes tokenizer configurations with different
compression comparable: a budget defined in real tokens would give the more compressive
tokenizer more optimizer steps for the same nominal budget. The position budget was calibrated
so that it corresponds to roughly $100$M real target-language tokens for the median
configuration. Because the padding fraction differs by tokenizer configuration and by
language, the real target-language token count of a given run is below its position budget and
is read from that run's own metadata rather than fixed in advance.
The seeds are $42$, $43$ and $44$, with $45$ and $46$ added in Hungarian under the adaptive
rule, in both model families. LoRA rank was ablated on Qwen2.5-0.5B: ranks $16$ and $64$, the
latter at $\alpha=128$, gave a minimal difference in downstream performance. The tied embedding
is created with \texttt{peft}'s \texttt{ensure\_weight\_tying}, and a runtime check compares
the storage of the input lookup and the output head and aborts if the two differ.

\paragraph{Parameters added by vocabulary extension.}
Under the \textsc{Vocabulary-Extension} configuration, 16,000 pieces are appended (ids 151,665 to 167,664, total vocabulary 167,665). The embedding matrix grows by 15,729 rows rather than 16,000: Qwen2.5-0.5B’s matrix has 151,936 rows for a 151,665-entry vocabulary, and the 271 unused rows are assigned to appended pieces first (151,936 + 15,729 = 167,665). The parameter count grows by 15,729 × 896 = 14.1M, where 896 is the embedding dimension and the embeddings are tied.

\section{Additional Natural Language Intrinsic Tokenization Results}\label{app:add_nl_intrinsic}
\subsection{Additional Intrinsic Metrics }\label{app:int_metrics}
\Cref{tab:app-sweep-recall} reports morphological alignment over $10$ tokenizers and $7$ languages: German, Finnish, Hungarian, Turkish, Hindi, Arabic and Tamil. \Cref{tab:app-surface-perlang} reports tokenization efficiency and information-theoretic metrics for German (deu), Finnish (fin), Hungarian (hun), Turkish (tur), Hindi (hin) and Swahili (swh). 
Notably, we see that 
R\'enyi-2.5 efficiency falls by $0.239$ ($95\%$ CI $[-0.281,-0.199]$) and bigram
entropy by $0.006$. These results suggest that LangMAP concentrates probability mass on the vocabulary subset that segments a
single language well.
On natural language, fertility changes are small and mixed in sign: it rises in three of the six languages (German $+0.0013$, Finnish $+0.0017$, Turkish $+0.0023$) and falls in the other three (Hungarian $-0.0024$, Hindi $-0.0062$, Swahili $-0.0038$), where a lower value is the improving direction.

\subsection{MorphScore measurement, Arabic gold segmentation and Hindi script coverage}
\label{app:morphscore-details}

\paragraph{ MorphScore computations.}
Each value in \cref{tab:morphscore} is measured over the word types of that language's Universal
Dependencies subset.  A value is marked in \cref{tab:morphscore} and in
\cref{tab:app-morphscore-full} when its $95\%$ confidence interval excludes zero under a
paired percentile bootstrap, that is, a procedure that draws $2{,}000$ resamples of the word
types with replacement, keeps each word type's two tokenizations together in every resample,
recomputes the difference on each resample, and reports the $2.5$th and $97.5$th percentiles
of the resulting $2{,}000$ differences. Under this procedure $36$ of the $40$ values of
\cref{tab:app-morphscore-full} are marked; the four that are not are the Llama-3.2-1B Arabic
recall and precision, the Llama-3.2-1B Hindi macro-F1, and the Llama-3.2-1B Hindi precision.
\Cref{tab:app-morphscore-full} shows all four MorphScore metrics for the ten rows.

\paragraph{Arabic gold segmentation.}
The Arabic gold segmentation was rebuilt for \cref{tab:morphscore} and
\cref{tab:app-sweep-recall}. 
The released prefix, stem and suffix split is an artifact of matching a vocalized lemma against an unvocalized word form: the mean stem length is $1.77$
characters and $26.5\%$ of word types have a one-character stem. 
We remove the lemma's
diacritics, normalize alef and ya, and locate the result inside the word form, which aligns
$16{,}014$ of the $19{,}126$ word types ($83.7\%$). Of the 16,014 aligned word types, 5,611 have a single gold morpheme. The scoring harness checks the token count before the morpheme count: a single-morpheme word type kept as one token is scored as correct, and a single-morpheme word type split into several tokens is excluded. The Qwen2.5-0.5B tokenizers keep 853 to 862 of the 5,611 single-morpheme word types whole and the Llama-3.2-1B tokenizers keep 200 to 211, which gives the retained counts 11,265 (Qwen Original), 11,256 (Qwen LangMAP), 10,614 (Llama Original) and 10,603 (Llama LangMAP). Table 1 prints the LangMAP count.
The $16.3\%$ that do not align are
non-concatenative forms, that is broken plurals, imperfective stem alternation and suppletion, so the Arabic values measure solely concatenative affixation on the alignable subset.

\begin{table}[h]
\centering\small\setlength{\tabcolsep}{4pt}
\adjustbox{max width=\columnwidth}{%
\begin{tabular}{@{}lrcccc@{}}
\toprule
\textbf{Lang.} & $n$ & $\Delta$\textbf{micro-F1} & $\Delta$\textbf{macro-F1} & $\Delta$\textbf{Rec.} & $\Delta$\textbf{Prec.} \\
\midrule
\multicolumn{6}{@{}l}{\emph{Qwen2.5-0.5B}} \\
Turkish              & 32{,}560 & $+0.0688^{\star}$ & $+0.0733^{\star}$ & $+0.2108^{\star}$ & $+0.0703^{\star}$ \\
Finnish              & 40{,}776 & $+0.0245^{\star}$ & $+0.0280^{\star}$ & $+0.1332^{\star}$ & $+0.0391^{\star}$ \\
Hungarian            &  7{,}924 & $+0.0350^{\star}$ & $+0.0405^{\star}$ & $+0.1530^{\star}$ & $+0.0385^{\star}$ \\
Arabic               & 11{,}256 & $+0.0743^{\star}$ & $+0.0775^{\star}$ & $+0.0663^{\star}$ & $+0.0350^{\star}$ \\
Hindi$^{\dagger}$    &  4{,}723 & $-0.0116^{\star}$ & $-0.0124^{\star}$ & $+0.1008^{\star}$ & $+0.0217^{\star}$ \\
\midrule
\multicolumn{6}{@{}l}{\emph{Llama-3.2-1B}} \\
Turkish              & 32{,}560 & $+0.0884^{\star}$ & $+0.0891^{\star}$ & $+0.1937^{\star}$ & $+0.0743^{\star}$ \\
Finnish              & 40{,}776 & $+0.0225^{\star}$ & $+0.0257^{\star}$ & $+0.1151^{\star}$ & $+0.0319^{\star}$ \\
Hungarian            &  7{,}924 & $+0.0413^{\star}$ & $+0.0473^{\star}$ & $+0.1659^{\star}$ & $+0.0440^{\star}$ \\
Arabic               & 10{,}603 & $+0.0074^{\star}$ & $+0.0078^{\star}$ & $-0.0043$         & $+0.0015$ \\
Hindi                &  4{,}723 & $-0.0009^{\star}$ & $-0.0005$         & $+0.0010^{\star}$ & $-0.0000$ \\
\bottomrule
\end{tabular}}
\caption{All four MorphScore metrics for the ten (model family, language) rows of
\cref{tab:morphscore}, which reports the first, third and fourth of them. Values are
\textsc{LangMAP} - \textsc{Original}; higher is better in all four columns.
$^{\star}$ marks a value whose $95\%$ confidence interval excludes zero under a paired
percentile bootstrap over word types with $2{,}000$ resamples, and $^{\dagger}$ the
Qwen2.5-0.5B Hindi row, whose recall is $1.0000$ by construction. Llama-3.2-1B Finnish has no downstream run.}
\label{tab:app-morphscore-full}
\end{table}

\subsection{Headroom Analysis}\label{app:headroom}
\begin{figure}
    \centering
    \includegraphics[width=\linewidth]{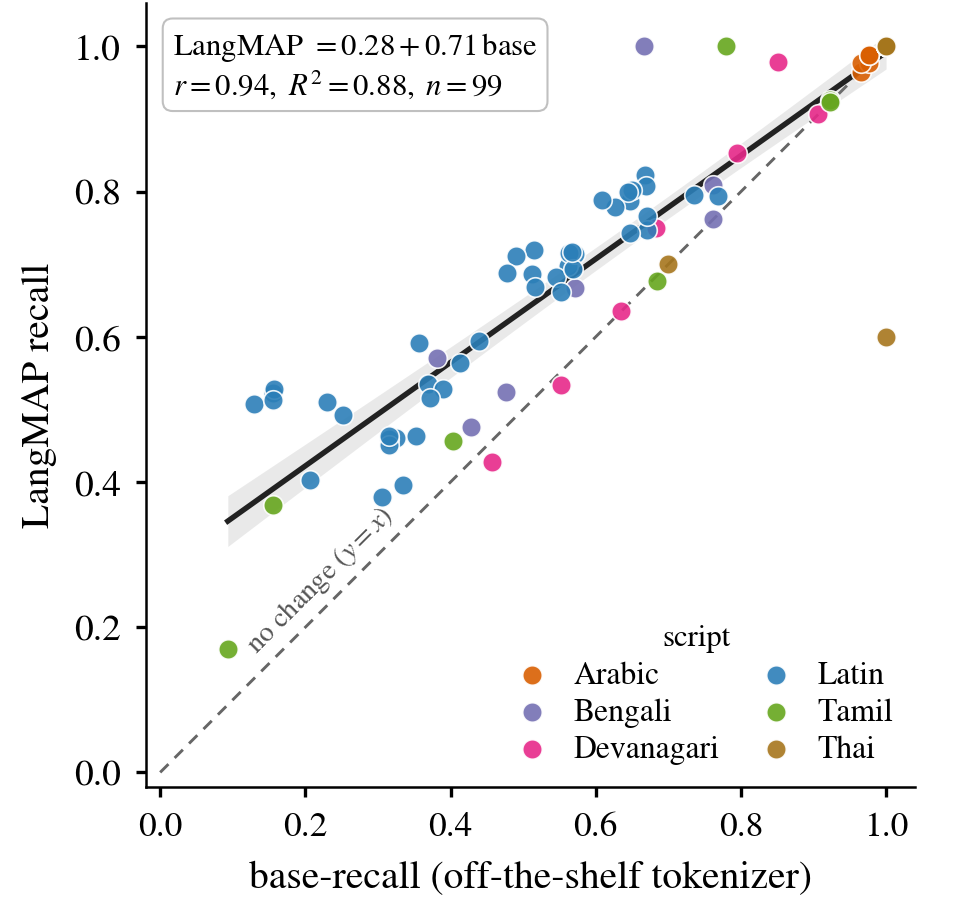}
    \caption{LangMAP MorphScore recall as a function of the original tokenizer's MorphScore recall. We see consistent improvements (points above diagonal) and that the gains are typically largest for tokenizers with the lowest base recall, i.e., where there is the most room for improvement. Our subsequent analysis shows that this result is contingent on the vocabulary containing sufficient coverage of a script/language. }
    \label{fig:headroom}
\end{figure}

\cref{fig:headroom} shows that LangMAP's MorphScore recall gain over the
base tokenizer is largest where the base tokenizer's recall is lowest, and near
zero where base recall is already high. This pattern shows the limits of what re-scoring can do. Concretely, LangMAP re-scores a fixed vocabulary and adds no tokens, so it changes only which of the existing pieces are used to segment a string.
Thus, a language can only benefit when the vocabulary already contains pieces that could yield a segmentation better aligned with that language's morpheme boundaries. What the vocabulary covers sets not only how large the change is but what kind of change it is: with enough pieces of the script, re-scoring selects among genuine alternatives, and with almost none, the character-level segmentation is the only alternative left for it to select.

We perform an additional analysis to explore this phenomenon. In  \cref{tab:deva-eng-subwords},  we count, for each of the eleven tokenizers, how many vocabulary entries are Devanagari pieces and how many are English or German pieces. 
\paragraph{How each column in \cref{tab:deva-eng-subwords} was computed.} A token's byte sequence is recovered
by reversing the GPT-2 byte-to-unicode map for byte-level tokenizers (Bloom,
Granite, Mistral~Nemo, Phi~4, Qwen\,2.5, Llama3), or by replacing the SentencePiece metaspace
marker with a space for SentencePiece tokenizers (Gemma~3, Mistral~7b, Phi~3,
XGLM, Yi). Added and special tokens are excluded. Before script classification, one optional leading space marker is stripped.
We compute the columns as follows; single-byte tokens are excluded from the counts throughout, so a count measures genuine merged subwords rather than the base byte alphabet. 
\begin{itemize}
    \item \textit{Devanagari, Complete}: tokens whose bytes decode as valid UTF-8 to a non-empty string in which every character
lies in the Devanagari block U+0900 to U+097F. A single Devanagari character is
three UTF-8 bytes, so single characters are included.
\item \textit{Devanagari, +Fragments}: the Complete count plus, for byte-level
tokenizers only, multi-byte tokens whose bytes are not valid standalone UTF-8 but
form a contiguous slice of a Devanagari UTF-8 sequence (lead byte 0xE0, second
byte 0xA4 or 0xA5, continuation bytes 0x80 to 0xBF). SentencePiece tokenizers
contribute no fragments because their pieces are whole characters, so the two
Devanagari columns are equal for them.
\item \textit{English subwords}: tokens that decode as valid UTF-8 to a non-empty string in which every character is an
ASCII letter a--z or A--Z. Digits, punctuation, mixed-script tokens, and
single-byte tokens (including lone letters) are excluded.
\item \textit{German}: tokens whose bytes decode
as valid UTF-8 to a non-empty string in which every character is a letter of the
German alphabet, that is a--z, A--Z, or one of \"a, \"o, \"u, \"A, \"O, \"U, \ss.
\end{itemize}

\Cref{tab:deva-eng-subwords} shows that the two scripts are covered unequally. Compared to German, Devanagari coverage is smaller and more variable, ranging from 3
pieces (Yi~1.5) to 16{,}189 (Bloom~3b), and falling below 40 for four tokenizers
(Yi~1.5: 3; Granite~3.0: 16; Phi-4: 23; Phi-3~mini: 39). These counts account for both ends of the trend. Because every tokenizer contains
tens of thousands of Latin pieces, German text has many alternative segmentations
under the vocabulary; LangMAP's per-language distribution selects among them, and
the selected segmentations align more closely with morpheme boundaries, which is evinced by German's mean recall gain of $+0.137$ over the ten tokenizers of \cref{tab:app-sweep-recall}.
Hindi is the other end, and the effect there is not that re-scoring leaves the
segmentation alone. A vocabulary with very few Devanagari pieces gives a Devanagari string
almost no segmentation other than the character-level one, which puts its MorphScore recall at $1.0000$ and makes that measurement uninformative
about morphological alignment. 
The eleven tokenizers of \cref{tab:deva-eng-subwords} span $3$ to
$16{,}189$ Devanagari pieces, which is why \cref{tab:app-sweep-recall} marks the two cases
separately instead of averaging over them. Of the nine tokenizers in that table's Hindi
column, four produce a recall difference that is a character-level ceiling effect, one
produces identical segmentations under both tokenizers, and the mean is taken over the four that
remain.

\begin{table*}[ht]
\centering
\small
\caption{Devanagari and Latin (English) subword inventories for the eleven subword
tokenizers we perform experiments with. For each tokenizer, the table reports how many vocabulary entries decode to complete Devanagari characters, how many additional byte-level fragments of Devanagari characters exist, and how many entries are Latin-letter
(English) subwords. The \emph{Vocabulary} column is the full vocabulary size, matching
\cref{tab:tokenizers}. The four subword counts are taken over vocabulary entries after
excluding added tokens, special tokens and single-byte tokens, so they do not sum to the
\emph{Vocabulary} column. }
\label{tab:deva-eng-subwords}
\begin{tabular}{l S[table-format=6.0] S[table-format=5.0] S[table-format=5.0] S[table-format=6.0] S[table-format=6.0]}
\toprule
 & & \multicolumn{2}{c}{Devanagari subwords} & \multicolumn{2}{c}{Latin-script subwords} \\
\cmidrule(lr){3-4} \cmidrule(lr){5-6}
{Tokenizer} & {Vocabulary} & {Complete} & {{+}Fragments} & {English} & {German} \\
\midrule
Bloom-3b       & 250880 & 16189 & 16210 & 89919  & 90100  \\
Gemma-3-1b & 262144 & 13684 & 13684 & 139481 & 141078 \\
Granite 3.0-8b & 49152  & 16    & 19    & 39727  & 39825  \\
Llama 3        & 128256 & 958   & 970   & 71150  & 71727  \\
Mistral 7b     & 32768  & 44    & 44    & 24952  & 25113  \\
Mistral Nemo   & 131072 & 1530  & 1542  & 70732  & 72196  \\
Phi 3 mini     & 32064  & 39    & 39    & 24016  & 24359  \\
Phi 4          & 100352 & 23    & 30    & 68643  & 68897  \\
Qwen\,2.5          & 151936 & 67    & 74    & 68871  & 69588  \\
XGLM 2.9b      & 256008 & 5667  & 5667  & 108841 & 112736 \\
Yi 1.5 6B      & 64000  & 3     & 3     & 38977  & 38986  \\
\bottomrule
\end{tabular}
\end{table*}

\begin{table*}[!ht]
    \centering
    \setlength{\tabcolsep}{4pt}\small
    \caption{\textbf{MorphScore recall difference, \textsc{LangMAP} - \textsc{Original},
    for $10$ tokenizers on $7$ languages.} 
    \textbf{Panel (a)}, Latin script. All
    $40$ differences are positive. 
    \textbf{Panel (b)}, non-Latin scripts, with two diagnostics beside each difference.
    \emph{Pieces} counts the vocabulary entries written entirely in that language's script,
    with the count of those longer than one character in parentheses. \emph{Char-split} is
    the frequency-weighted fraction of word types cut at every character position, for the
    original tokenizer and then for LangMAP. $^{\dagger}$ marks the $6$ measurements in which the two tokenizers produce the same segmentation of every word type, producing a difference of $0$. 
    $\ddagger$ marks the 7 measurements in which the recall difference is a character-level ceiling effect; see App. B.3. 
    The $14$ measurements with neither mark show  differences ranging from $-0.005$
    to $+0.119$. The mean row of panel (b) uses only these $14$, so it is not an average over all families.
    Yi-1.5-6b appears in panel (a) only.  Yi's vocabulary contains $3$ Devanagari pieces and $0$ Tamil pieces and does not have byte fallback.
    Those two scripts must be represented primarily by unknown tokens, and the analysis is confounded.}
    \label{tab:app-sweep-recall}
    \adjustbox{max width=\textwidth}{%
    \begin{tabular}{@{}lccccc@{}}
        \toprule
        \multicolumn{6}{@{}l}{\emph{(a) Latin-script languages}} \\
        \textbf{Tokenizer} & German & Finnish & Hungarian & Turkish & \textbf{Mean} \\
        \midrule
Bloom-3b          & $+0.180$ & $+0.089$ & $+0.104$ & $+0.151$ & $+0.131$ \\
Gemma-3-1b        & $+0.126$ & $+0.123$ & $+0.130$ & $+0.167$ & $+0.137$ \\
Granite 3.0-8b    & $+0.134$ & $+0.107$ & $+0.171$ & $+0.118$ & $+0.132$ \\
Llama-3.2-1b      & $+0.179$ & $+0.115$ & $+0.166$ & $+0.194$ & $+0.164$ \\
Mistral-7b-v0.3   & $+0.090$ & $+0.117$ & $+0.151$ & $+0.131$ & $+0.122$ \\
Mistral-Nemo-12b  & $+0.107$ & $+0.099$ & $+0.142$ & $+0.149$ & $+0.124$ \\
Phi-4             & $+0.177$ & $+0.127$ & $+0.151$ & $+0.175$ & $+0.157$ \\
Qwen2.5-0.5B      & $+0.179$ & $+0.133$ & $+0.153$ & $+0.211$ & $+0.169$ \\
XGLM-2.9b         & $+0.011$ & $+0.042$ & $+0.053$ & $+0.092$ & $+0.050$ \\
Yi-1.5-6b         & $+0.185$ & $+0.105$ & $+0.148$ & $+0.144$ & $+0.146$ \\
        \midrule
\textit{Mean} & $\mathit{+0.137}$ & $\mathit{+0.106}$ & $\mathit{+0.137}$ & $\mathit{+0.153}$ & $\mathit{+0.133}$ \\
        \bottomrule
    \end{tabular}}

    \vspace{8pt}

    \adjustbox{max width=\textwidth}{%
    \begin{tabular}{@{}lccccccccc@{}}
        \toprule
        \multicolumn{10}{@{}l}{\emph{(b) Non-Latin-script languages}} \\
        & \multicolumn{3}{c}{Hindi (Devanagari)} & \multicolumn{3}{c}{Arabic} & \multicolumn{3}{c}{Tamil} \\
        \cmidrule(lr){2-4}\cmidrule(lr){5-7}\cmidrule(lr){8-10}
        \textbf{Tokenizer} & $\Delta$Rec. & Pieces & Char-split & $\Delta$Rec. & Pieces & Char-split & $\Delta$Rec. & Pieces & Char-split \\
        \midrule
Bloom-3b          & $+0.030$ & 16{,}189 (16{,}038) & $0.002{\to}0.002$ & $+0.040$ & 20{,}587 (20{,}394) & $0.001{\to}0.001$ & $+0.119$ & 6{,}194 (6{,}114) & $0.000{\to}0.000$ \\
Gemma-3-1b        & $+0.008$ & 13{,}684 (13{,}519) & $0.002{\to}0.002$ & $-0.005$ & 8{,}408 (8{,}008) & $0.002{\to}0.002$ & $+0.085$ & 3{,}850 (3{,}769) & $0.001{\to}0.001$ \\
Granite 3.0-8b    & $+0.000^{\dagger}$ & 16 (0) & $1.000{\to}1.000$ & $+0.002^{\ddagger}$ & 79 (25) & $0.157{\to}0.181$ & $+0.000^{\dagger}$ & 25 (0) & $1.000{\to}1.000$ \\
Llama-3.2-1b      & $+0.001^{\ddagger}$ & 958 (845) & $0.365{\to}0.366$ & $-0.004$ & 3{,}729 (3{,}594) & $0.002{\to}0.002$ & $+0.000^{\dagger}$ & 3 (0) & $0.101{\to}0.101$ \\
Mistral-7b-v0.3   & $0.000^{\ddagger}$ & 44 (0) & $1.000{\to}0.998$ & $+0.000^{\dagger}$ & 71 (2) & $0.298{\to}0.298$ & $+0.000^{\dagger}$ & 22 (0) & $1.000{\to}0.989$ \\
Mistral-Nemo-12b  & $+0.031$ & 1{,}530 (1{,}412) & $0.070{\to}0.097$ & $+0.005$ & 9{,}350 (9{,}197) & $0.001{\to}0.001$ & $+0.052$ & 525 (466) & $0.013{\to}0.013$ \\
Phi-4             & $+0.041^{\ddagger}$ & 23 (1) & $0.613{\to}0.807$ & $+0.001^{\ddagger}$ & 105 (36) & $0.164{\to}0.169$ & $+0.000^{\dagger}$ & 3 (0) & $0.101{\to}0.101$ \\
Qwen2.5-0.5B      & $+0.101^{\ddagger}$ & 67 (1) & $0.613{\to}0.984$ & $+0.066$ & 4{,}008 (3{,}449) & $0.001{\to}0.001$ & $+0.321^{\ddagger}$ & 28 (0) & $0.101{\to}1.000$ \\
XGLM-2.9b         & $+0.007$ & 5{,}667 (5{,}536) & $0.000{\to}0.000$ & $+0.018$ & 13{,}168 (12{,}962) & $0.001{\to}0.001$ & $-0.002$ & 1{,}945 (1{,}876) & $0.000{\to}0.000$ \\
        \midrule
\textit{Mean} & $\mathit{+0.019}$ & $n{=}4$ &  & $\mathit{+0.020}$ & $n{=}6$ &  & $\mathit{+0.064}$ & $n{=}4$ &  \\
        \bottomrule
    \end{tabular}}
\end{table*}\clara{figure out $^{\ddagger}$!!}

\begin{table*}[!ht]
    \centering
    \setlength{\tabcolsep}{4pt}\footnotesize
    \caption{\textbf{Surface efficiency and information-theoretic metrics,
    per language.} $\Delta=\textsc{LangMAP}-\textsc{Original}$ (raw,
    per-checkpoint convention). Improving direction is \emph{lower} for
    fertility and \emph{higher} for the other three; small brackets are
    $95\%$ CIs and $^{\star}$ marks a CI excluding $0$.  Compression rate is shown $\times10^{3}$.}
    \label{tab:app-surface-perlang}
    \adjustbox{max width=\textwidth}{
    \begin{tabular}{@{}lccccc@{}}
        \toprule
        \textbf{Lang} & $n$ & $\Delta$\textbf{Fertility} & $\Delta$\textbf{Compr.}\,($\times10^{3}$) & $\Delta$\textbf{Bigram }$H$ & $\Delta$\textbf{R\'enyi-2.5} \\
        \midrule
        deu & 10 & $+0.0013^\star$ {\scriptsize$[+0.0002,+0.0023]$} & $+0.11$ {\scriptsize$[0.00,+0.19]$} & $-0.0068^\star$ {\scriptsize$[-0.0085,-0.0050]$} & $-0.251^\star$ {\scriptsize$[-0.317,-0.175]$} \\
        fin & 10 & $+0.0017$ {\scriptsize$[-0.0032,+0.0056]$} & $+0.11$ {\scriptsize$[-0.13,+0.33]$} & $+0.0010$ {\scriptsize$[-0.0044,+0.0056]$} & $-0.343^\star$ {\scriptsize$[-0.464,-0.240]$} \\
        hun & 10 & $-0.0024^\star$ {\scriptsize$[-0.0049,0.0000]$} & $-0.09$ {\scriptsize$[-0.22,+0.05]$} & $-0.0078^\star$ {\scriptsize$[-0.0103,-0.0051]$} & $-0.303^\star$ {\scriptsize$[-0.420,-0.193]$} \\
        tur & 10 & $+0.0023$ {\scriptsize$[-0.0046,+0.0087]$} & $+0.38$ {\scriptsize$[-0.07,+0.93]$} & $-0.0012$ {\scriptsize$[-0.0065,+0.0036]$} & $-0.109^\star$ {\scriptsize$[-0.176,-0.037]$} \\
        hin & 10 & $-0.0062$ {\scriptsize$[-0.0185,+0.0003]$} & $-0.10$ {\scriptsize$[-0.34,+0.06]$} & $-0.0086$ {\scriptsize$[-0.0226,+0.0002]$} & $-0.154$ {\scriptsize$[-0.398,+0.003]$} \\
        swh & 10 & $-0.0038^\star$ {\scriptsize$[-0.0069,-0.0010]$} & $-0.16$ {\scriptsize$[-0.32,+0.00]$} & $-0.0103^\star$ {\scriptsize$[-0.0140,-0.0064]$} & $-0.605^\star$ {\scriptsize$[-0.778,-0.407]$} \\
        \bottomrule
    \end{tabular}}
\end{table*}

\section{Additional Code Intrinsic Tokenization Results}\label{app:add_code_intrinsic}

The values used in \cref{tab:code-summary} are shown in \cref{tab:code-overall-grid}. A value is marked in \cref{tab:code-summary} when its $95\%$
confidence interval excludes zero under the percentile bootstrap  procedure described in \cref{app:exp_setup}.

\crnew{%
\begin{table*}[t]
\centering\footnotesize\setlength{\tabcolsep}{4.5pt}
\caption{Per-(model, language) difference in \emph{overall} AST-leaf-boundary alignment,
LangMAP minus the model's own tokenizer, for every one of the $8\times9=72$ (model,
programming language) measurements.
Alignment is the fraction of leaf-node spans in the parse tree whose start and end both
coincide with a token boundary; higher is better. All $72$ differences are positive, from
$+0.003$ to $+0.066$, with a mean of $+0.027$. The right column and bottom row give
per-model and per-language means.
The evaluation set is $100$ source files per language, taken in parquet row order from the StarCoder eval split. }
\label{tab:code-overall-grid}
\adjustbox{max width=\textwidth}{
\begin{tabular}{@{}lccccccccc|c@{}}
\toprule
\textbf{Model} & C & C\texttt{++} & C\# & Go & Java & JS & PHP & Py & TS & \textbf{Mean} \\
\midrule
BLOOM-3B           & $+0.052$ & $+0.036$ & $+0.033$ & $+0.028$ & $+0.029$ & $+0.060$ & $+0.039$ & $+0.018$ & $+0.041$ & $\mathbf{+0.037}$ \\
Gemma\,2-2B        & $+0.066$ & $+0.018$ & $+0.010$ & $+0.021$ & $+0.011$ & $+0.027$ & $+0.012$ & $+0.017$ & $+0.016$ & $\mathbf{+0.022}$ \\
Granite\,3.0-8B    & $+0.028$ & $+0.018$ & $+0.009$ & $+0.022$ & $+0.011$ & $+0.026$ & $+0.012$ & $+0.019$ & $+0.011$ & $\mathbf{+0.017}$ \\
Llama\,3.2-1B      & $+0.049$ & $+0.030$ & $+0.021$ & $+0.038$ & $+0.040$ & $+0.035$ & $+0.016$ & $+0.037$ & $+0.030$ & $\mathbf{+0.033}$ \\
Mistral-NeMo-12B   & $+0.046$ & $+0.026$ & $+0.030$ & $+0.040$ & $+0.041$ & $+0.035$ & $+0.014$ & $+0.043$ & $+0.028$ & $\mathbf{+0.034}$ \\
Phi-4              & $+0.049$ & $+0.030$ & $+0.022$ & $+0.038$ & $+0.042$ & $+0.035$ & $+0.016$ & $+0.037$ & $+0.030$ & $\mathbf{+0.033}$ \\
Qwen\,2.5-0.5B     & $+0.049$ & $+0.030$ & $+0.022$ & $+0.037$ & $+0.042$ & $+0.035$ & $+0.016$ & $+0.036$ & $+0.030$ & $\mathbf{+0.033}$ \\
XGLM-2.9B          & $+0.023$ & $+0.006$ & $+0.009$ & $+0.015$ & $+0.003$ & $+0.012$ & $+0.003$ & $+0.008$ & $+0.006$ & $\mathbf{+0.009}$ \\
\textit{Mean} & $\mathit{+0.045}$ & $\mathit{+0.024}$ & $\mathit{+0.019}$ & $\mathit{+0.030}$ & $\mathit{+0.027}$ & $\mathit{+0.033}$ & $\mathit{+0.016}$ & $\mathit{+0.027}$ & $\mathit{+0.024}$ & $\mathit{+0.027}$ \\
\bottomrule
\end{tabular}}
\end{table*}}

\section{Additional Downstream Results}
\label{app:downstream}

\paragraph{Per-language MultiBLiMP accuracy for jointly-adapted model.}
\cref{tab:f15-joint} shows per-language MultiBLiMP accuracy for one model adapted to all five languages jointly. 
With 3 seeds a percentile bootstrap over seeds returns the smallest and the largest of the
three observed differences, so we report the spread as the observed range over the 3 seeds.
The three per-seed differences on Qwen2.5-0.5B are Turkish $+0.0098$, $+0.0052$, $+0.0103$;
Finnish $+0.0241$, $+0.0023$, $+0.0019$; Hungarian $+0.0095$, $+0.0012$, $+0.0189$; and
Arabic and Hindi change sign across seeds. On Llama-3.2-1B they are Turkish $-0.0218$,
$-0.0293$, $-0.0235$; Finnish $+0.0047$, $-0.0062$, $+0.0237$; Hungarian $+0.0083$,
$+0.0059$, $-0.0059$; Arabic $+0.0041$, $-0.0008$, $-0.0058$; and Hindi $+0.0000$, $-0.0055$,
$-0.0007$.

\crnew{%
\begin{table*}[t]
\centering\small
\setlength{\tabcolsep}{4pt}
\begin{tabular}{@{}l ccccc@{}}
\toprule
 & Original & Base-Unigram & LangMAP & Placebo & Vocab.-ext. \\
\midrule
\multicolumn{6}{@{}l}{\emph{Target-language FLORES+ bits per byte} (lower is better)} \\
Finnish   & 1.3849 & 1.4278 & 1.4245 & n/a & \textbf{1.2745} \\
Hungarian & 1.3293 & 1.3514 & 1.3494 & 1.3607 & \textbf{1.2069} \\
Turkish   & 1.2922 & 1.3108 & 1.3097 & n/a & \textbf{1.2417} \\
Arabic    & 0.9335 & 0.9413 & 0.9394 & 0.9500 & \textbf{0.9324} \\
Hindi     & 0.6526 & 0.6592 & 0.6591 & 0.6587 & \textbf{0.5946} \\
\midrule
\multicolumn{6}{@{}l}{\emph{English FLORES+ bits per byte} (lower is better)} \\
Finnish   & 1.0819 & 1.0895 & 1.0907 & n/a & 1.1673 \\
Hungarian & 1.0746 & 1.0820 & 1.0815 & 1.0826 & 1.1060 \\
Turkish   & 1.0726 & 1.0809 & 1.0816 & n/a & 1.1244 \\
Arabic    & 1.0370 & 1.0458 & 1.0442 & 1.0451 & 1.0484 \\
Hindi     & 1.0504 & 1.0579 & 1.0565 & 1.0569 & 1.0564 \\
\midrule
\multicolumn{6}{@{}l}{\emph{English Belebele accuracy, pooled over languages and seeds} (higher is better)} \\
Pooled    & 0.4993 & 0.4932 & 0.4981 & 0.5000 & 0.4807 \\
\bottomrule
\end{tabular}
\caption{FLORES+ bits per byte and pooled English Belebele accuracy after adapting Qwen2.5-0.5B to one language at a time. Each bits-per-byte value is the mean over 3 seeds and \textbf{bold} marks the lowest value in a row; the two bits-per-byte panels and the Belebele row run in opposite directions, as marked. \textsc{Placebo} was run in Arabic, Hindi and Hungarian only. The Belebele row is a mean over languages and seeds, with no bootstrap and no correction applied. }
\label{tab:f15-bpb}
\end{table*}}

\begin{table*}[t]
\centering\small
\setlength{\tabcolsep}{6pt}
\begin{tabular}{@{}l ccc c@{}}
\toprule
Language & Original & LangMAP (routed) & LangMAP $-$ Original & Routed $-$ Oracle \\
\midrule
\multicolumn{5}{@{}l}{\emph{Qwen2.5-0.5B}} \\
Turkish   & 0.7907 & 0.7991 & $+0.0084$ & $0.0000$ \\
Finnish   & 0.7843 & 0.7938 & $+0.0095$ & $0.0000$ \\
Hungarian & 0.8923 & 0.9022 & $+0.0099$ & $0.0000$ \\
Arabic    & 0.9155 & 0.9158 & $+0.0003$ & $0.0000$ \\
Hindi     & 0.9330 & 0.9307 & $-0.0023$ & $0.0000$ \\
\midrule
\multicolumn{5}{@{}l}{\emph{Llama-3.2-1B}} \\
Turkish   & 0.8576 & 0.8328 & $-0.0249$ & $0.0000$ \\
Finnish   & 0.8355 & 0.8429 & $+0.0074$ & $0.0000$ \\
Hungarian & 0.9254 & 0.9282 & $+0.0028$ & $0.0000$ \\
Arabic    & 0.8936 & 0.8927 & $-0.0008$ & $0.0000$ \\
Hindi     & 0.9666 & 0.9645 & $-0.0021$ & $0.0000$ \\
\bottomrule
\end{tabular}
\caption{Per-language MultiBLiMP accuracy for one model adapted to all five languages jointly, in each of the two model families, means over 3 seeds; higher is better. \emph{Routed} selects the segmentation at inference without a language label; \emph{oracle} selects it using the true language label. The routed and oracle accuracies are identical in all $30$ (model family, language, seed) combinations, and on FLORES+ each family's tokenizer pool assigns $997$ of $997$ sentences to the correct language's distribution in each of the five languages. Absolute accuracies are below the single-language values of \cref{tab:f15-multiblimp} (Turkish $0.7907$ here against $0.8387$ there for Qwen2.5-0.5B); we attribute that difference to one model covering five languages rather than one. The differences in the third column are computed over 3 seeds rather than over evaluation items, for the reason and with the per-seed values given in \cref{app:downstream}.}
\label{tab:f15-joint}
\end{table*}

\paragraph{Bits per byte on Llama-3.2-1B.}
\Cref{tab:f15-llama-bpb} reports FLORES+ bits per byte for Llama-3.2-1B adapted to one
language at a time, under the protocol of \cref{tab:f15-multiblimp}. The original tokenizer has
the lower value in all four target languages and in English.

\crnew{%
\begin{table}[h]
\centering\small
\setlength{\tabcolsep}{5pt}
\begin{tabular}{@{}l cc c@{}}
\toprule
Language & Original & LangMAP & LangMAP $-$ Original \\
\midrule
\multicolumn{4}{@{}l}{\emph{Target-language FLORES+ bits per byte} (lower is better)} \\
Turkish   & 1.1746 & 1.2094 & $+0.0348$ \\
Hungarian & 1.1539 & 1.2053 & $+0.0514$ \\
Arabic    & 0.9118 & 0.9274 & $+0.0156$ \\
Hindi     & 0.5699 & 0.5736 & $+0.0036$ \\
\midrule
\multicolumn{4}{@{}l}{\emph{English FLORES+ bits per byte} (lower is better)} \\
Turkish   & 0.9822 & 0.9926 & $+0.0103$ \\
Hungarian & 0.9825 & 0.9925 & $+0.0100$ \\
Arabic    & 0.9781 & 0.9812 & $+0.0031$ \\
Hindi     & 0.9738 & 0.9759 & $+0.0021$ \\
\bottomrule
\end{tabular}
\caption{FLORES+ bits per byte after adapting Llama-3.2-1B to one language at a time, under the
protocol of \cref{tab:f15-multiblimp}; lower is better in both panels. Finnish was not run in
this family, and this family was run with the original tokenizer and LangMAP only. Each value
is the mean over 3 seeds, with seed standard deviations of $0.0001$ to $0.0012$. English
Belebele is not tabulated for this family: the pretrained model scores $0.347$ against a
four-way chance level of $0.25$, so it does not measure English retention at this scale.}
\label{tab:f15-llama-bpb}
\end{table}}

\paragraph{Confidence intervals and $p$-value computations.}

A paired bootstrap over evaluation items with seeds as replicates is computed alongside each difference from the original tokenizer in \cref{tab:f15-multiblimp}. Significance is only marked when the confidence interval for that difference excludes 0 after multiple testing adjustment: we use the Holm procedure at $\alpha=0.05$, over the $18$ measurements taken in the
Qwen2.5-0.5B panel and the $4$ in the Llama-3.2-1B panel.
The Hungarian runs, except Base-Unigram, were extended to 5 seeds under the adaptive rule of \cref{sec:adaptation}.

\paragraph{Validation bits per byte for jointly-trained model.}
Bits per byte on the in-training validation split of the model trained jointly on five languages is $1.0772$ for the
original tokenizer and $1.0942$ for LangMAP on Qwen2.5-0.5B, aggregated over the five languages
and 3 seeds, a difference of $+0.017$ in the same direction as the single-language setting.

\paragraph{MorphScore differences and MultiBLiMP differences.}
We tested, across the nine (language, model family) combinations we adapted, whether MorphScore differences (LangMAP minus the original tokenizer) are positively associated with the MultiBLiMP differences. 
Each downstream difference is paired with the MorphScore difference of the tokenizer that run was trained with, taken from \cref{tab:morphscore}.
The prediction is not supported. The Spearman correlation is negative under every MorphScore column, at $-0.27$ on recall, $-0.35$ on precision and $-0.20$ on each of the two F1 columns of \cref{tab:app-morphscore-full}. 
The intrinsic and downstream differences agree in sign in $4$ to $6$ of the nine measurements, but an exact sign test shows this result is not distinguishable from an at-chance result. 
Turkish has the largest MorphScore recall difference in both model families, $+0.211$ on Qwen2.5-0.5B and $+0.194$ on Llama-3.2-1B, together with downstream differences of $-0.008$ and $-0.011$, in the opposite direction to the prediction. 

\paragraph{LangMAP convergence.} \label{ap:convergence}

\Cref{tab:convergence} shows that LangMAP estimation has low sample complexity. 
Across German, Finnish and Turkish, re-estimating the language-specific distribution from only $1{,}000$ documents gives a held-out fertility within $0.054\%$ of the reference estimate, averaged over the three languages. 
The last row shows the scale at which every LangMAP distribution in this paper was fit, $25$M tokens per language.

\begin{table}[t]
    \centering\small
  \caption{\textbf{Convergence of the language-specific distribution.} Each value is
    the difference in fertility between a distribution $\phi_\ell$ re-estimated from the first $N$ documents of that language's corpus and the reference distribution,
    written as a percentage of the reference value. Fertility is  computed on the FLORES-200 devtest set.
    The base vocabulary is Qwen2.5-0.5B's and is held fixed; only the per-language distribution is re-estimated. The reference is the $N{=}32{,}000$ fit.
    The last row is the scale at which every LangMAP distribution in this paper was fit: $25$M tokens per language.  }
    \label{tab:convergence}
    \begin{tabular}{@{}rccc@{}}
        \toprule
        \textbf{Documents} ($N$) & German & Finnish & Turkish \\
        \midrule
        $125$      & $0.306\%$ & $0.187\%$ & $0.230\%$ \\
        $500$      & $0.067\%$ & $0.181\%$ & $0.040\%$ \\
        $1{,}000$  & $0.002\%$ & $0.101\%$ & $0.059\%$ \\
        $4{,}000$  & $0.026\%$ & $0.059\%$ & $0.076\%$ \\
        $16{,}000$ & $0.005\%$ & $0.047\%$ & $0.016\%$ \\
        $32{,}000$ & \multicolumn{3}{c}{\textit{reference}} \\
        \midrule
        $25$M tokens & $0.009\%$ & $0.009\%$ & $0.011\%$ \\
        \bottomrule
    \end{tabular}
\end{table}

\paragraph{Label-free segmentation selection.}
\label{app:routing}

We investigate whether the correct language's segmentation is used at inference time in the LangMAP tokenization scheme. We evaluate on $12$ languages (German, Finnish, Hungarian, Turkish, Hindi, Bengali, Tamil, Arabic, Thai, which are used for our MorphScore evaluation and the label-free routing measurement; and  Basque, Korean, Russian). 
Here the input is truncated to $W$ leading words and the base model is \texttt{EuroLLM-1.7B}. 
For $12$ languages ($400$ sentences each), LangMAP chooses the \unilm segmentation from the input's ground-truth language with $100\%$ accuracy once $8$ words of the sequence are given as input, for all $12$ languages and for the Latin-script subset alike (Table~\ref{tab:routing}). When the input is at least $8$ words, routing is done with perfect accuracy. 
Note that this is a different measurement from the routing result mentioned in \cref{sec:extrinsic-results}. 

\begin{table}[t]
    \centering\small
    \caption{\textbf{Label-free language routing accuracy.} As a function of input length (first $W$ whitespace words), the fraction of FLORES inputs that were segmented according to the ``correct'' language's unigram distribution via LangMAP's
    maximum-likelihood selection rule; ``full'' $=$ whole sentence. Base $=$ EuroLLM-1.7B,
    $400$ sentences/language. ``Latin-5'' restricts both the input languages and LangMAP's language set to the five Latin-script languages.}
    \label{tab:routing}
    \adjustbox{max width=\columnwidth}{
    \begin{tabular}{@{}lcccccc@{}}
        \toprule
        Input ($W$ words) & $1$ & $2$ & $3$ & $5$ & $8$ & full \\
        \midrule
        All $12$ languages & $.888$ & $.961$ & $.988$ & $.998$ & $1.00$ & $1.00$ \\
        Latin-$5$ only     & $.803$ & $.922$ & $.976$ & $.995$ & $1.00$ & $1.00$ \\
        \bottomrule
    \end{tabular}}
\end{table}

\onecolumn
\section{Further Details on the UnigramLM Algorithm}\label{app:uni_background}

This appendix provides additional detail on the \unilm algorithm summarized in \cref{sec:unigramlm}. We give the full inference and EM derivations (\cref{sec:inference-derivation,sec:em-motivation}, respectively) and the vocabulary pruning procedure (\cref{sec:vocab-pruning}). These derivations are largely reproduced from \citet{meister2025unigramlm}; we refer the reader to this resource for a more detailed presentation of the algorithm.

\subsection{Inference Derivation}\label{sec:inference-derivation}

For reference, the deterministic relationship between token sequences and strings under the \unilm generative model is:
\begin{equation} \label{eq:prob_deterministic_string}
    P(\stringrv = \str \mid \tokenseqrv = \tokens) = 
    \left\{\begin{array}{lr}
        1 & \texttt{if } \str = \detokfunc(\tokens) \\
        0 & \texttt{otherwise}
    \end{array}\right.
\end{equation}
Recall from the main text that MAP inference seeks:
\begin{equation}\tag{\ref{eq:map_inference}}
    \bestsegmentation \defeq \argmax_{\tokens \in \vocab^*} 
    \post{\tokens}{\str}
\end{equation}
Using the relationships in 
\cref{eq:prob_token_sequence,eq:prob_deterministic_string}, 
we can rewrite \cref{eq:map_inference} as the following equivalent optimization problem:
\begin{subequations}
\begin{align}
\bestsegmentation 
&\defeq \argmax_{\tokens \in \vocab^*} 
    \post{\tokens}{\str} 
    \tag{\ref{eq:map_inference}} \\
&=\argmax_{\tokens \in \allsegmentations_{\vocab}(\str)} 
    \frac{\ptokenseq{\tokens}}
    {\sum_{\tokens' \in \allsegmentations_{\vocab}(\str)} 
    \ptokenseq{\tokens'}} \label{step:1}\\
&= \argmax_{\tokens \in \allsegmentations_{\vocab}(\str)} 
    \ptokenseq{\tokens} \label{step:2}\\
&= \argmax_{\tokens \in \allsegmentations_{\vocab}(\str)} 
    \prod_{t=1}^{|\tokens|} \unigramdistparams[\token_t] 
    \label{eq:viterbi}
\end{align}
\end{subequations}
where we first restrict the considered sequences of tokens to $\allsegmentations_{\vocab}(\str)$ (\ref{step:1}) since other sequences have probability zero given 
\cref{eq:prob_deterministic_string}, and apply Bayes' rule, noting that $P(\stringrv = \str \mid \tokenseqrv = \tokens; 
\unigramdistparams) = 1$ for all 
$\tokens \in \allsegmentations_{\vocab}(\str)$.
Then, we drop the denominator (\ref{step:2}), since it does not depend on $\tokens$ and thus does not affect the $\argmax$. 
Finally, we apply the definition in 
\cref{eq:prob_token_sequence} (\ref{eq:viterbi}).

\subsection{Application of EM for Learning $\unigramdistparams$}\label{sec:em-motivation}

As stated in \cref{sec:unigramlm}, \unilm learns $\unigramdistparams$ by approximately maximizing the observed data log-likelihood via EM. Here we provide the full logic behind why this is a valid approach.

Under the \unilm generative process, a ``complete'' dataset consists of $(\str, \tokens)$ pairs, i.e., strings and the token sequences that produced them. Denoting such a dataset as 
$\mathcal{X} = \{(\str_m, \tokens_m)\}_{m=1}^M$, its 
log-likelihood is:
\begin{equation}\label{eq:complete-likelihood}
    \likelihood(\mathcal{X}; \unigramdistparams) \defeq 
    \sum_{m=1}^M \log\joint{\str_m}{\tokens_m}
\end{equation}
which we refer to as the \emph{complete} data log-likelihood. 
Crucially, this quantity cannot be optimized directly, since the token sequences $\tokens_m$ are latent: we only observe the detokenized strings $\str = \detokfunc(\tokens)$, not the segmentations that produced them. We thus instead turn to 
maximizing the \emph{observed} data log-likelihood:
\begin{subequations}\label{eq:corpus-likelihood}
\begin{align}
    \likelihood(\corpus; \unigramdistparams)
    &\defeq
    \sum_{m=1}^M \log\pstring{\str_m}\\
    &\;=\;
    \sum_{m=1}^M \log \sum_{\tokens \in 
    \allsegmentations_{\vocab}(\str_m)} \ptokenseq{\tokens}
\end{align}
\end{subequations}
where $\corpus = \{\str_m\}_{m=1}^M$ is the corpus of strings. 
However, this value is difficult to optimize directly due to the log-sum structure. This is where the expectation-maximization (EM) algorithm comes in.

By Jensen's inequality, we can show that the \emph{expected value} of the complete log-likelihood serves as a lower bound for the \emph{observed} log-likelihood. We provide this derivation specifically for our setting. Let $\Q(\unigramdistparams;\unigramdistparamscur) $  denote the expected complete data log-likelihood:
\begin{equation}
    \Q(\unigramdistparams;\unigramdistparamscur) \defeq \sum_{m=1}^M\underset{\tokens\sim \tokenseqrv\,\mid\, \stringrv=\str_m;\unigramdistparamscur}{\E}\big[\log \joint{\str_m}{\tokens}\big]
\end{equation}

For any valid probability distribution  $P(\tokenseqrv=\tokens)$, given our established relationships between $P(\tokenseqrv=\tokens)$ and $ \pstring{\str}$, we have by Jensen's inequality that:
\begin{align*}
\log \pstring{\str}
&= \log \sum_{\tokens\in \allsegmentations_{\vocab}(\str)} P(\tokenseqrv=\tokens)\,\frac{\ptokenseq{\tokens}}{P(\tokenseqrv=\tokens)}\\
&\ge \sum_{\tokens\in \allsegmentations_{\vocab}(\str)} P(\tokenseqrv=\tokens)\,\log \frac{\ptokenseq{\tokens}}{P(\tokenseqrv=\tokens)}\label{eq:jensen}
\end{align*}
Let's now choose $P(\tokenseqrv=\tokens)$ to be \(\postcur{\tokens}{ \str}\), the posterior under our current parameter beliefs for a fixed $\str$. 
Applying Jensen's inequality (above) to our definition of the observed data log-likelihood from \cref{eq:corpus-likelihood} gives us
\begin{subequations}
\begin{align}
\likelihood&(\corpus;\unigramdistparams)= \sum_{m=1}^M \log\sum_{\tokens\in\allsegmentations_{\vocab}(\str_m)} \ptokenseq{\tokens}\\
&\ge \sum_{m=1}^M\sum_{\tokens\in \allsegmentations_{\vocab}(\str_m)}\postcur{\tokens}{\str_m}
\big[\log \ptokenseq{\tokens}-\log \postcur{\tokens}{\str_m}\big]\label{eq:apply-jensens}\\
&= \sum_{m=1}^M\sum_{\tokens\in \allsegmentations_{\vocab}(\str_m)}\postcur{\tokens}{\str_m}
\log \joint{\str_m}{\tokens} \label{eq:subbing-post} \\
& \qquad\qquad\qquad\qquad-\sum_{m=1}^M\sum_{\tokens\in \allsegmentations_{\vocab}(\str_m)}\postcur{\tokens}{\str_m}\log \postcur{\tokens}{\str_m}
\nonumber\\
&= \underbrace{\sum_{m=1}^M\underset{\tokens\sim \tokenseqrv\,\mid\, \stringrv=\str_m;\unigramdistparamscur}{\E}\big[\log \joint{\str_m}{\tokens}\big]}_{\Q(\unigramdistparams;\unigramdistparamscur)} \label{eq:elbo-Q-H} \\
& \qquad\qquad\qquad\qquad+ \sum_{m=1}^M\underbrace{-\underset{\tokens\sim \tokenseqrv\,\mid\, \stringrv=\str_m;\unigramdistparamscur}{\E}\big[\log \postcur{\tokens}{\str_m}\big]}_{\Hent\big(\tokenseqrv\,\mid \,\stringrv=\str_m;\unigramdistparamscur\big)} \nonumber
\end{align}
\end{subequations}
When going to \cref{eq:subbing-post}, we use the fact that for any $\tokens \in \allsegmentations_{\vocab}(\str_m)$, it must be that  $\joint{\str_m}{\tokens} = \ptokenseq{\tokens}$. \cref{eq:elbo-Q-H} simply inserts some common definitions. Namely, the definitions of expected values and (Shannon) entropy.
The entropy term $\Hent\big(\tokenseqrv\,\mid \,\stringrv=\str_m;\unigramdistparamscur\big)$ is constant with respect to $\unigramdistparams$, and so we can ignore it in our optimization.

The EM algorithm uses the above relationship and proposes a procedure for iteratively maximizing the expected complete data log-likelihood with respect to model parameters, where the expectation is taken with respect to the posterior distribution defined by current parameter beliefs. 
The E-step computes this expected value; the M-step maximizes it with respect to the free parameters. In our case, expected token counts are a sufficient statistic for the M-step objective, leading to 
the algorithm given in \cref{sec:unigramlm}.

This procedure can be intuitively understood as iteratively performing maximum likelihood estimation for a categorical model, where observed token counts are replaced by posterior-expected counts induced by the distribution over latent segmentations given current parameter beliefs. The usage of EM for this problem is taken from prior work and its theoretical validity for problems of the same structure has 
been well established 
\cite{em_categorical,brown-etal-1993-mathematics}.

\subsection{\unilm Vocabulary Pruning}\label{sec:vocab-pruning}
As our method keeps $\vocab$ fixed, we omit this step entirely; we include this description of pruning done in the original \unilm algorithm only for completeness.
The \unilm algorithm was designed for the case that we do not start with a known, fixed $\vocab$. It thus incorporates a strategy for choosing $\vocab$ by initializing $\vocab$ as an over-sized set and adding a pruning step at the end of each EM iteration. This pruning step uses current parameter beliefs to estimate the log-likelihood contribution of each token to the expected observed data log-likelihood. It then prunes a percentage of tokens that contribute least, stopping once the desired vocabulary size (a hyperparameter set by the user) has been reached.
\citet{land2026piecesdoesunigramtokenization} discuss this strategy and its effects. Our algorithm uses several of the methods from \unilm but is not tied to any particular method for choosing $\vocab$: the vocabulary may equally be one learned by BPE, or that of a pretrained model.

% \subsection{Expected Token Counts as a Sufficient Statistic}\label{sec:exp_tok_count_proof}
% \Clara{TODO}

%%%%%%%%%%%%%%%%%%%%%%%%%%%%%%%%%%%%%%%%%%%%%%%%%%%%%%%%%%%%%%%%%%%%%%%%%%%%%%%
%%%%%%%%%%%%%%%%%%%%%%%%%%%%%%%%%%%%%%%%%%%%%%%%%%%%%%%%%%%%%%%%%%%%%%%%%%%%%%%

\end{document}